\documentclass{ieeeaccess}
\usepackage{amsmath,amssymb,amsfonts}
\usepackage{algorithmic}
\usepackage{graphicx,multirow}
\usepackage[font={sf,scriptsize},           
            labelfont={bf,color=accessblue},
            caption=false]{subfig} 
\usepackage{textcomp}
\definecolor{assumpH1}{RGB}{237, 125, 49}
\definecolor{assumpH2}{RGB}{113, 174, 71}
\definecolor{assumpH3}{RGB}{0, 176, 240}
\definecolor{assumpH41}{RGB}{112, 48, 160}
\definecolor{assumpH42}{RGB}{255, 0, 0}
\definecolor{assumpH5}{RGB}{255, 51, 204}

\begin{document}
\history{Under Review: Date of publication xxxx 00, 0000, date of current version xxxx 00, 0000.}
\doi{10.1109/ACCESS.2022.0122113}

\title{Motion Capture Benchmark of Real Industrial Tasks and Traditional Crafts for Human Movement Analysis}
\author{\uppercase{Brenda Elizabeth Olivas-Padilla}\authorrefmark{1}, \uppercase{Alina Glushkova}\authorrefmark{1}, and \uppercase{Sotiris Manitsaris}\authorrefmark{1}}

\address[1]{The Centre for Robotics, Mines Paris, Université PSL, 75006 Paris, France (e-mail: name.lastname@minesparis.psl.eu)}

\tfootnote{The research leading to these results has received funding from the CARNOT Projet Fédérateur "Usine responsable" and the Horizon 2020 Research and Innovation Programme under Grant Agreement No. 820767, CoLLaboratE project, and Grant No. 822336, Mingei project.}

\markboth
{Brenda Elizabeth Olivas-Padilla \headeretal: Preparation of Papers for IEEE TRANSACTIONS and JOURNALS}
{Brenda Elizabeth Olivas-Padilla \headeretal: Preparation of Papers for IEEE TRANSACTIONS and JOURNALS}

\corresp{Corresponding author: Brenda Elizabeth Olivas-Padilla (e-mail: brenda.olivas@ minesparis.psl.eu).}

\begin{abstract}
Human movement analysis is a key area of research in robotics, biomechanics, and data science. It encompasses tracking, posture estimation, and movement synthesis. While numerous methodologies have evolved over time, a systematic and quantitative evaluation of these approaches using verifiable ground truth data of three-dimensional human movement is still required to define the current state of the art. This paper presents seven datasets recorded using inertial-based motion capture. The datasets contain professional gestures carried out by industrial operators and skilled craftsmen performed in real conditions in-situ. The datasets were created with the intention of being used for research in human motion modeling, analysis, and generation. The protocols for data collection are described in detail, and a preliminary analysis of the collected data is provided as a benchmark. The Gesture Operational Model, a hybrid stochastic-biomechanical approach based on kinematic descriptors, is utilized to model the dynamics of the experts' movements and create mathematical representations of their motion trajectories for analysis and quantifying their body dexterity. The models allowed accurate the generation of human professional poses and an intuitive description of how body joints cooperate and change over time through the performance of the task.
\end{abstract}


\titlepgskip=-21pt

\maketitle

\section{Introduction}
\label{sec:introduction}
Previous studies of human motion data helped researchers better comprehend body dynamics and their stochastic behavior. Capturing raw data from human movement performed in different contexts permits quantifying and a better understanding of the motion parameters as well as the factors that impact motor performance. 
By analyzing this data, hidden parameters can be revealed, useful for motion evaluation in sports, rehabilitation, and arts but also in more professional and industrial contexts for ergonomic monitoring. Professional diseases linked to ergonomy, such as Muscular Skeletal Disorders (MSDs), constitute an important issue causing negative effects not only on the operators' health but also on the productivity of a factory/workshop.

In such context, numerous motion capture initiatives have been undertaken in various fields. Most of them have been made publicly available for visualization and analysis, and they include data corresponding to different activities of human everyday life covering from ample body motions to very fine facial expressions. These datasets can be categorized based on the technologies used (marker-based or marker-less motion capture, etc.), on the activities recorded (everyday activity, sports, etc.), or on the number of users (single user vs. multiuser interaction). For example, HumanEva and MoVi \cite{Sigal2010, Ghorbani2021} are two existing datasets that contain video and marker-based motion capture (MoCap) data of a single person performing ordinary activities (like walking and jogging) and sports motions (boxing). General body movements have also been recorded with a monocular camera in the HMDB51 \cite{Kuehne2011} dataset, including user-object interaction, human-to-human interaction, and facial expressions. A multimodal human action database MHAD \cite{Ofli2013} has been published, including limb actions recorded with a camera, accelerometers, and a microphone. Another more recent initiative has been done in the CMU dataset \cite{CMU}, also including the multimodal signal from food preparation activities. For motions in multiperson interactions and scenarios, van der Aa et al. \cite{VanDerAa2011} presented the UMPM benchmark. Also, the KIT dataset recorded human-to-human interaction activities while manipulating various objects \cite{Mandery2015}.

In all the aforementioned studies, everyday activity has monopolized the interest of researchers performing motion capture. However, in the last decade, it has become more and more interesting to use motion capture and to apply data analysis methods to scenarios inspired by a professional context where human operators perform their tasks. The table below presents recent works of recording industry-oriented human motion data. Several examples can be found in the construction industry since it is one of the most affected by intense physical activity.

\begin{table*}[!htbp] 
\centering
\caption{Datasets available and employed by the community.}
\begin{tabular}{llllll} 
\hline
\multicolumn{1}{c}{\textbf{ Dataset }} & \multicolumn{1}{c}{\textbf{ Technology used }}                                                                                               & \multicolumn{1}{c}{\textbf{ Activity recorded }}                                                                                     & \multicolumn{1}{c}{\begin{tabular}[c]{@{}c@{}}\textbf{ Year of }\\\textbf{publication }\end{tabular}} & \multicolumn{1}{c}{\begin{tabular}[c]{@{}c@{}}\textbf{ \# of }\\\textbf{subjects }\end{tabular}} & \multicolumn{1}{c}{\begin{tabular}[c]{@{}c@{}}\textbf{ Type of Environment }\\\textbf{the data has been }\\\textbf{captured in }\end{tabular}}  \\ 
\hline\hline
DeTECLoad ~                            & Single IMU                                                                                                                                   & \begin{tabular}[c]{@{}l@{}}Construction workers \\load carrying tasks\end{tabular}                                                   & 2020                                                                                                  & 14                                                                                               & \begin{tabular}[c]{@{}l@{}}Controlled conditions \\in a laboratory\end{tabular}                                                                 \\ 
\hline
AnDy project ~                         & \begin{tabular}[c]{@{}l@{}}Marker based motion\\capture \\system \\4 pressure sensors \\2 RGB cameras \\~Full body IMUs + glove\end{tabular} & \begin{tabular}[c]{@{}l@{}}Industry oriented activities \\(screw high/middle/low,\\untie knot etc.)\end{tabular}                     & 2020                                                                                                  & 13                                                                                               & \begin{tabular}[c]{@{}l@{}}Controlled conditions \\in a laboratory\end{tabular}                                                                 \\ 
\hline
VV-Conlot ~                            & \begin{tabular}[c]{@{}l@{}}RGB monocular\\camera \\3 IMUs\end{tabular}                                                                       & \begin{tabular}[c]{@{}l@{}}Construction industry oriented \\(painting, vacuum cleaning, \\jumping from the stairs etc.)\end{tabular} & 2021                                                                                                  & 13                                                                                               & \begin{tabular}[c]{@{}l@{}}Controlled conditions \\in a laboratory\end{tabular}                                                                 \\ 
\hline
IKEA ASM                               & 3 RGB cameras                                                                                                                                & \begin{tabular}[c]{@{}l@{}}Furniture assembling \\(screwing etc.)\end{tabular}                                                       & 2020                                                                                                  & 48                                                                                               & \begin{tabular}[c]{@{}l@{}}Controlled conditions \\in a laboratory\end{tabular}                                                                 \\ 
\hline
WGD ~                                  & \begin{tabular}[c]{@{}l@{}}Marker based motion \\capture system with \\8 cameras\end{tabular}                                                & \begin{tabular}[c]{@{}l@{}}Assembly line working gestures \\(hammering, screwing etc.)\end{tabular}                                  & 2021                                                                                                  & 8                                                                                                & \begin{tabular}[c]{@{}l@{}}Controlled conditions \\in a laboratory\end{tabular}                                                                 \\
\hline\hline
\end{tabular}
\end{table*}

To detect excessive load-carrying tasks, Lee et al. \cite{Lee2020} have focused on creating a dataset with a non-invasive single IMU sensor. The recorded data has served to automatically predict load-carrying weights and postures using CNNs. Fourteen subjects were recorded performing six different carrying modes. The dataset's analysis consists of modeling, classification, and predicting load-carrying weights. Another interesting dataset was recorded in the framework of the AnDy EU project \cite{Maurice2019}, where various sensors were used, such as a full-body IMU suit including a glove for finger motion, a marker-based motion capture system, a finger pressure sensor, and 2 video cameras. The subjects performed industry-oriented activities inspired by car manufacturing. The data was annotated and labeled and is intended for use by researchers developing algorithms for classifying, predicting, or evaluating human movement in industrial settings. The evaluation focuses mostly on label reliability, not movement analysis itself. The VTT-Conlot dataset includes motion data inspired by the construction industry recorded with 3 IMUs, with 13 subjects \cite{Makela2021}. The principal goal of this dataset is to be used for activity recognition and classification. Its evaluation refers to sensor location, modalities used, and features extracted. However, contrary to previous examples cited, the VTT-ConIot validated and compared its data also with real unannotated data belonging to real workers in a real construction site (the real data is not included in the VTT-Conlot dataset). The IKEA ASM dataset is a multi-view, furniture assembly video dataset that includes depth, atomic actions, object segmentation, and human pose \cite{Ben-Shabat2020}. One of the particularities of this one is that it includes unusual human poses performed while assembling furniture, but it does not include any IMU-captured data and aims mostly at solving computer vision challenges. The WGD dataset provides data recorded with a marker-based system of subjects performing assembly line working activities \cite{Tamantini2021}. A kinematic evaluation of the data has been performed, showing that the dataset can be used for human ergonomics evaluations.

All the aforementioned works went beyond recording everyday activities and focused on professional tasks/gestures/postures. However, there is still a need for MoCap data that include a greater diversity of movements, particularly professional gestures captured in real-world scenarios. Most of the datasets available were recorded inside a laboratory, causing approximate measures since they may lack authenticity and are not real workplace scenarios. Thus, this paper presents datasets created to capture and study operators' and artisans' gestures in their professional settings and real environment, performed under real conditions. 

\begin{table*}[!htbp] 
\centering
\caption{Overview of the generated datasets.}
\begin{tabular}{llll} 
\hline
\multicolumn{1}{c}{\textbf{ Dataset }}                                        & \multicolumn{1}{c}{\textbf{ Activity recorded }}                                                     & \multicolumn{1}{c}{\textbf{ Location }}                                                                                              & \multicolumn{1}{c}{\begin{tabular}[c]{@{}c@{}}\textbf{ \# of }\\\textbf{subjects }\end{tabular}}  \\ 
\hline\hline
TV assembly                                                                   & \begin{tabular}[c]{@{}l@{}}Drilling, connecting components on \\a production line, etc.\end{tabular} & Turkey, Arcelik factory                                                                                                              & 5                                                                                                 \\ 
\hline
Airplane floater assebly                                                      & \begin{tabular}[c]{@{}l@{}}Hammering the rivet, placing the \\bucking bar, etc.\end{tabular}         & Romania, Romaero factory                                                                                                             & 2                                                                                                 \\ 
\hline
Silk weaving                                                                  & \begin{tabular}[c]{@{}l@{}}Jacquard weaving gestures with \\looms of different sizes\end{tabular}    & Germany, Krefeld silk museum                                                                                                         & 2                                                                                                 \\ 
\hline
Glass blowing                                                                 & \begin{tabular}[c]{@{}l@{}}Shaping the decanter, blowing through \\the blowpipe, etc.\end{tabular}   & France, Cerfav, glass blowing workshop                                                                                               & 1                                                                                                 \\ 
\hline
Mastic cultivation                                                            & \begin{tabular}[c]{@{}l@{}}Sweeping the soil, embroidering \\the tree, etc.\end{tabular}             & \begin{tabular}[c]{@{}l@{}}Greece, Chios island, \\Mastic museum fields (outdoor) and \\in controlled indoor conditions\end{tabular} & 2                                                                                                 \\ 
\hline
\begin{tabular}[c]{@{}l@{}}Postures according to \\EAWS protocol\end{tabular} & Bending, rotating the torso, etc.                                                                    & Controlled conditions in a laboratory                                                                                                & 10                                                                                                \\
\hline\hline
\end{tabular}
\end{table*}

The recording procedures and processing methods are detailed in this paper. Additionally, it is provided a first analysis of the seven datasets using an analytical model called the Gesture Operational Model (GOM), which was proposed in a previous work \cite{Manitsaris2020}. In this analysis are created interpretable motion representations based on GOM that can be used to artificially generate human movements and explain the inter-collaboration of joints during the performance of the modeled movements. The results comprise the forecasting performance measures on every dataset and a dexterity analysis of professional tasks. The dexterity analysis applies GOM's mathematical representations to describe the performance of professional gestures. Dexterity can be defined as the skill to perform a given movement or task using the hands or other body parts. In addition, a method for identifying the most significant joint motion descriptors for modeling and recognizing a set of human movements is described. This knowledge can then be utilized to determine the ideal sensor configuration for human motion recognition problems.

\section{Data acquisition}\label{ap_dataaq}
This section begins with a description of the MoCap system used for recording, followed by information on the subjects and gestures captured for each dataset. 
\subsection{Motion capture technology}
The BioMed bundle motion capture system from Nansense Inc.\footnote{Baranger Studios, Los Angeles, CA, USA} was utilized to capture the gestures of industrial operators and craftsmen. The system is composed of a full-body suit with 52 IMUs strategically positioned across the torso, limbs, and hands. At a rate of 90 frames per second, the sensors measure the orientation and acceleration of body segments on the articulated spine chain, shoulders, arms, legs, and fingertips. After a recording, the Euler local joint angles on the X, Y, and Z axes are automatically calculated through the Nansense Studio's inverse kinematics solver and stored in a Biovision Hierarchy format (BVH). A BVH file is a text file comprised of two parts. The first part provides a hierarchical description of the skeleton, beginning with the root (hips) and proceeding to the extremities of each limb. The second part of the file contains, for each frame of the recording, the absolute position of the root of the skeleton and the angles of the joints defined in the first part of the BVH file.

\subsection{Subjects recruited}
For the creation of each dataset, industrial operators and skilled artisans consented to be recorded in their actual workplace while wearing the Nansense suit in accordance with the General Data Protection Regulation (GDPR) principles. Firstly, industrial operators from a television plant in Istanbul, Turkey, and an aerospace company in Bucharest, Romania, were captured as they carried out their professional tasks. Four healthy people, three men and one woman, participated in the MoCap recording session at the television plant. Their average age was 31.5±6.2 years, their height was 167.8±4.6 cm, and their average weight of 65.3±9.9 kg. Two male subjects participated in the MoCap session for the recordings in the aerospace company. They had an average age of 50±5 years, a height of 170±2 cm, and a weight of 77±1.4 kg.

Ten healthy individuals consented to participate in MoCap recordings of potentially dangerous ergonomic postures in a neutral environment laboratory. The subjects consisted of three women and seven men. The average age was 28.7±4.6 years, with an average height of 172.9±9.2 cm, and the average weight was 70.5±12.9 kg. None of them sustained musculoskeletal injuries, and they all completed all trials in under one hour.

Gestures of skilled artisans performing three different crafts were recorded. The first is a master silk weaver recorded at a traditional jacquard workshop in Krefeld, Germany. The expert's height was 168 cm, and his weight was 62 kg. The second artisan is a master glassblower who was recorded in action during a glassblowing workshop. The glassblower's height was 177 cm, and his weight was 73 kg. Finally, two mastic farmers were recorded at a mastic cultivation field in Chios, Greece. Their average age was 30.5±5.5 years, their height was 178.8±8.5 cm, and their average weight was 69.3±8.0 kg.

\subsection{Recording of the professional tasks} 
The procedure followed for each recording is outlined next, as well as a description of each captured task. Before recording, a calibration procedure was done. The subject assumed different postures, such as I-pose or T-pose, and performed different movements, like walking or touching his fingertips, each for 10 seconds. In order to facilitate the later annotation and segmentation of the data, only operators and artisans were asked to explain each component of the task prior to the recording.

\subsubsection{Industrial-related tasks}\label{secP2}
The gestures performed in two industrial settings have been recorded, delivering natural movements while operators execute industrial tasks. The tasks were captured on-site during regular production by actual operators. 

\begin{figure*}[!htbp] 
\centering
\subfloat[$\text{TVA}_1$]{\includegraphics[width=0.15\textwidth, height=30mm]{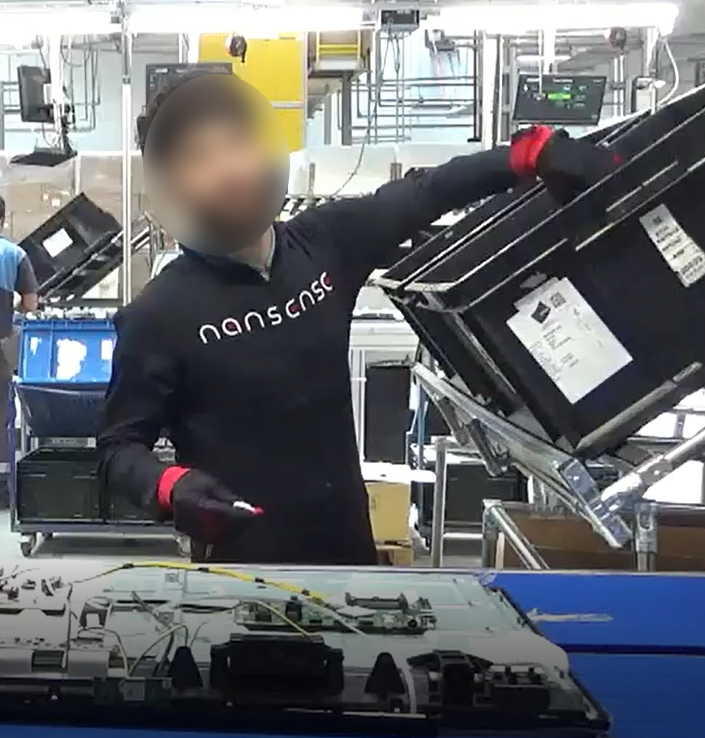}\label{fig:GV2a}}
\hfil
\subfloat[$\text{TVA}_2$]{\includegraphics[width=0.15\textwidth, height=30mm]{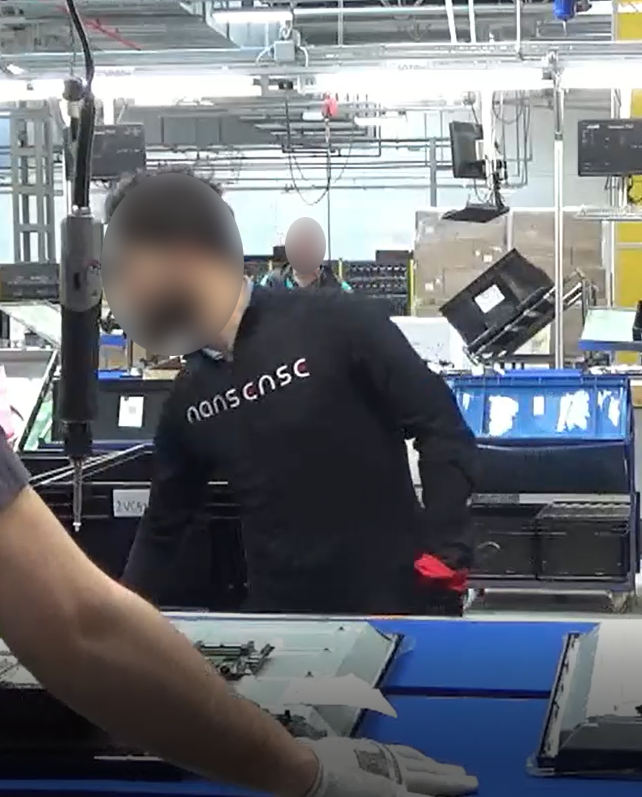}}
\hfil
\subfloat[$\text{TVA}_3$]{\includegraphics[width=0.15\textwidth, height=30mm]{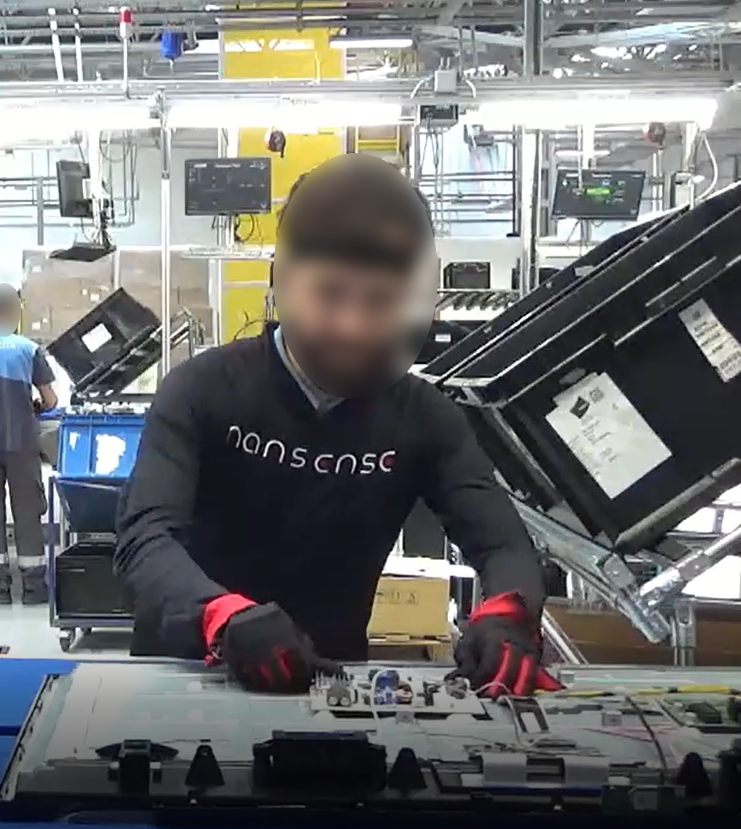}}
\hfil
\subfloat[$\text{TVA}_4$]{\includegraphics[width=0.15\textwidth, height=30mm]{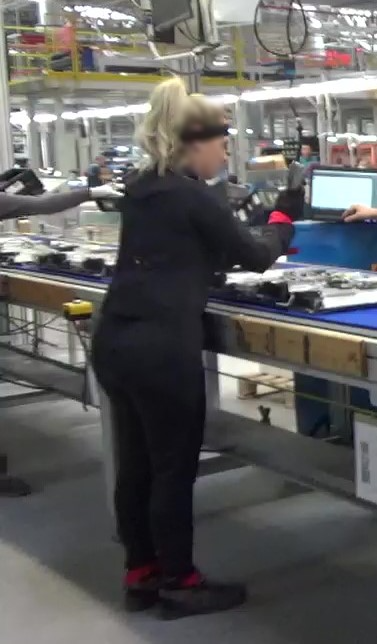}}
\hfil
\subfloat[$\text{TVP}_8$]{\includegraphics[width=0.15\textwidth, height=30mm]{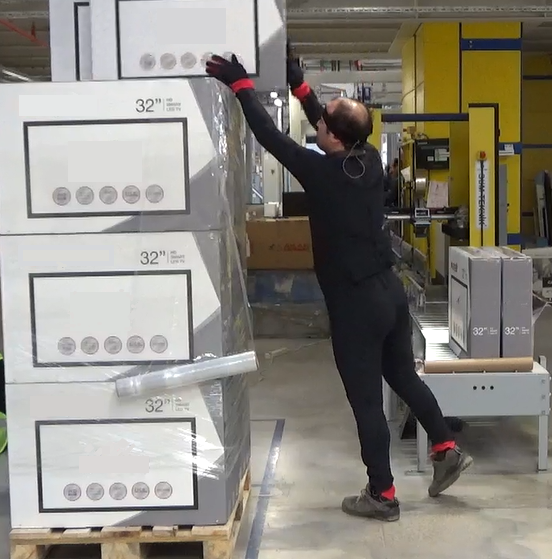}\label{fig:GV2d}}
\caption{Professional gestures in television manufacturing.}
\label{fig:GV2}
\end{figure*}

\paragraph{Television manufacturing}
Two tasks were recorded at a television manufacturing plant related to assembly and packaging. The set of gestures involved in each task is designated by the abbreviations \textbf{TVA} (assembly) and \textbf{TVP} (packaging). Fig. \ref{fig:GV2} illustrates some of the gestures recorded in television assembly and packaging.

The television assembly task consists of mounting electronic circuit boards to a television chassis and using a power tool to drive screws into the boards to secure them firmly. For this task it was defined the following gesture vocabulary:
\begin{itemize}
\item $\text{TVA}_1$: Reaching high with one hand, above shoulder level, to pick one component (circuit board) from a container.
\item $\text{TVA}_2$: Reaching low with the other empty hand, below the knee level, to pick up the second component (wire) from a second container.
\item $\text{TVA}_3$: Connecting the components and placing the board on the chassis to be screwed.
\item $\text{TVA}_4$: Drilling four screws on the circuit board by holding the driller with the right hand and placing the screws with the left.
\end{itemize}

The final operation required stacking the completed, boxed televisions on wooden pallets and wrapping them in a plastic membrane for shipping (TVP). The following set of gestures were recorded for this task:
\begin{itemize}
\item $\text{TVP}_1$: Placing eight TVs on a wooden pallet (bottom level).
\item $\text{TVP}_2$: Preparing to wrap the bottom level with a membrane.
\item $\text{TVP}_3$: Wrapping the bottom level.
\item $\text{TVP}_4$: Placing eight TVs on top of the bottom level (second level).
\item $\text{TVP}_5$: Wrapping the second level with a plastic membrane.
\item $\text{TVP}_6$: Placing eight TVs on top of the second level (third level).
\item $\text{TVP}_7$: Wrapping the third level with a plastic membrane.
\item $\text{TVP}_8$: Placing eight TVs on top of the third level (fourth level).
\item $\text{TVP}_9$: Wrapping the fourth level with a plastic membrane.
\end{itemize}
Boxes are given to the operator through a conveyor belt. He places one box at a time onto the pallet using both hands. After stacking eight boxes on a single level, he grabs the plastic membrane with both hands and wraps them by going around them with it. After wrapping them properly, the operator proceeds to stack boxes on top of the previous one wrapped, repeating the process. The task is complete when there are four levels of boxes on the pallet.

All tasks associated with television assembly were recorded over the course of an eight-hour shift, with one subject recorded installing the circuit boards during the first half of the shift and another recorded drilling the circuit boards to the television chassis during the second half. Three subjects were recorded separately for the packaging tasks during one shift. 

\paragraph{Airplane floater assembly}
The complete riveting task for an airplane floater was captured in an aerospace company. The floater is a plane component that enables planes to float when they land on water. The set of gestures recorded from this task is denoted as \textbf{APA}. Collaboration between two operators is essential for this activity. Therefore, their data were collected sequentially; one person wore the MoCap suit to capture their movement while collaborating and then donned it to the second person and continued the activity. As a result, the following gestures were recorded, which are also illustrated in Fig. \ref{fig:GV3}:
\begin{figure*}[!htbp] 
\centering
\subfloat[$\text{APA}_1$]{\includegraphics[width=0.2\textwidth, height=30mm]{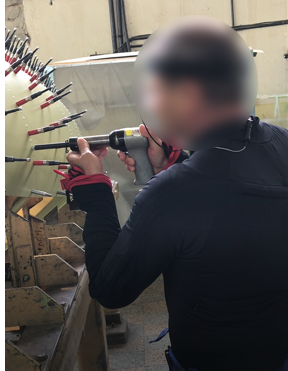}\label{fig:GV3a}}
\hfil
\subfloat[$\text{APA}_2$]{\includegraphics[width=0.2\textwidth, height=30mm]{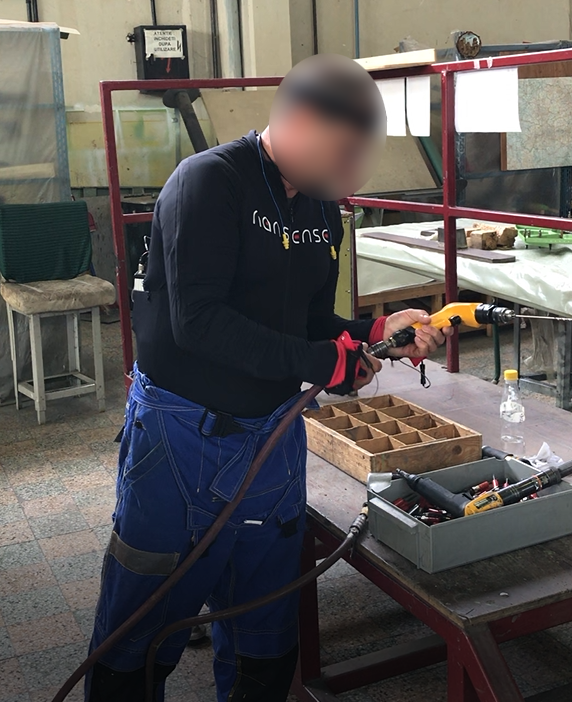}\label{fig:GV3b}}
\hfil
\subfloat[$\text{APA}_3$]{\includegraphics[width=0.2\textwidth, height=30mm]{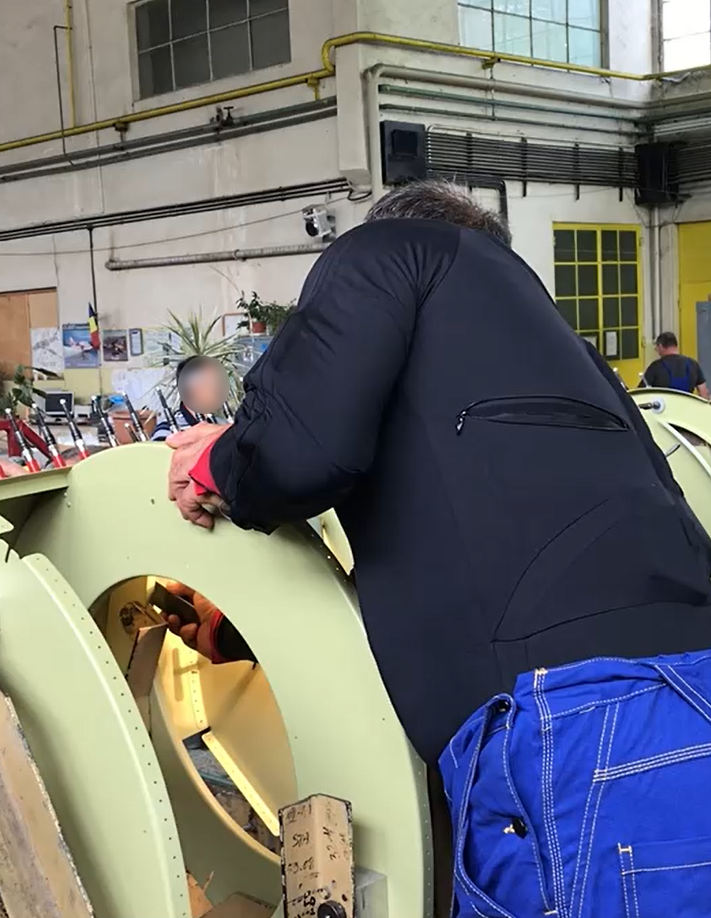}\label{fig:GV3c}}
\caption{Example of airplane assembly gestures.}
\label{fig:GV3}
\end{figure*}

\begin{itemize}
\item $\text{APA}_1$: Rivet with the pneumatic hammer.
\item $\text{APA}_2$: Prepare the pneumatic hammer and grab rivets.
\item $\text{APA}_3$: Place the bucking bar to counteract the incoming rivet.
\end{itemize}

One iteration of rivet assembly consisted of the first operator placing a rivet in one hole (Fig. \ref{fig:GV3a}). The second operator from the opposite side of the floater then positions the bucking bar to counter the rivet (Fig. \ref{fig:GV3c}). After precisely positioning the bucking bar, the second operator signals the first operator to activate the pneumatic hammer. The first operator verifies the proper placement of the assembled rivet by touching it, then moves on to the next hole and the process is repeated. After completing one line of rivets, the first operator grabs additional rivets and prepares the pneumatic hammer for the second line (Fig. \ref{fig:GV3b}).

The movement of the fingers during the riveting with the pneumatic hammer was not recorded because the operator could not work realistically while wearing the MoCap gloves. The operator needed to touch with his bare hands the rivet to determine whether it was positioned correctly.

\paragraph{Postures with varying ergonomic risk level}
A recording protocol was designed to capture 28 postures with varying ergonomic risk levels based on the European Assembly Worksheet (EAWS) \cite{Schaub2013}.

\begin{figure*}[!htbp]
\centering
    \subfloat[$\text{ERGD}_7$]{\includegraphics[width=0.2\textwidth, height=30mm]{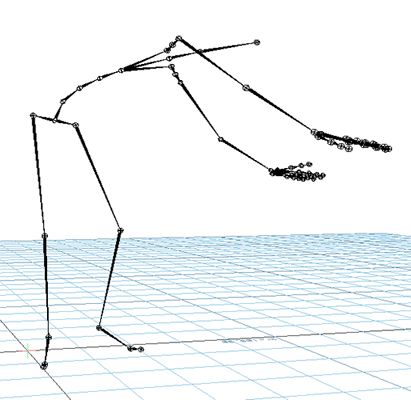}}
    \hfil
    \subfloat[$\text{ERGD}_{19}$]{\includegraphics[width=0.2\textwidth, height=30mm]{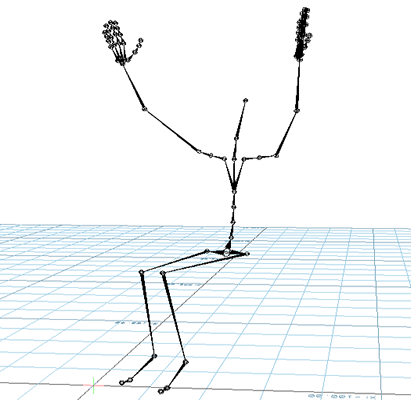}}
    \hfil
    \subfloat[$\text{ERGD}_{28}$]{\includegraphics[width=0.2\textwidth, height=30mm]{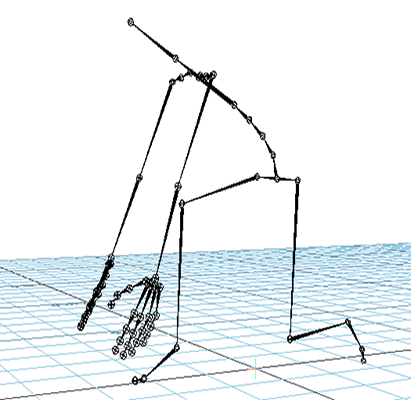}}
\caption{Example of postures contained in ERGD.}
\label{fig:GVE}
\end{figure*}

Each posture was repeated three times, giving a total of 84 MoCap recordings per subject. The recorded postures were neutral as they were not associated with a specific activity but rather served solely to demonstrate several ergonomically incorrect postures. The postures can be divided into three main categories: those performed standing, those performed seated on a chair, and those executed while kneeling. The postures are progressing from comfortable postures to increasingly more uncomfortable but never dangerous ones. All postures were held for six seconds, and no particular discomfort was reported. This set of 28 postures with different ergonomic risk levels is denoted as \textbf{ERGD}. Three postures assumed by the subjects are shown in Fig. \ref{fig:GVE}.
Initially, the subject is standing with a straightened back. The subject then assumes the following three postures:
\begin{itemize}
\item $\text{ERGD}_1$: The subject remains standing straight up, with the arms relaxed (I-pose).
\item $\text{ERGD}_2$: The subject rotates their torso to the left as far as they can for six seconds.
\item $\text{ERGD}_3$: The subject bends laterally the torso to the left for six seconds.
\end{itemize}
For the next three postures, the torso is slightly bent forwards:
\begin{itemize}
\item $\text{ERGD}_4$: The subject remains in the bending position for six seconds.
\item $\text{ERGD}_5$: While the subject is bending forward, they rotate their torso to the left and hold this position for six seconds.
\item $\text{ERGD}_6$: While the subject bends forward and rotates their torso to the left, they extend their arm as if trying to reach something that is on the ground.
\end{itemize}
The next three postures have the torso bending forward at a large angle ($>60^{\circ}$):
\begin{itemize}
\item $\text{ERGD}_7$: The subject remains in the bending position for six seconds.
\item $\text{ERGD}_8$: While the subject has bent forwards, they rotate their torso to the left and hold this position for six seconds.
\item $\text{ERGD}_9$: While the subject bends forward and rotates their torso to the left, they extend their arm as if trying to reach something that is on the ground.
\end{itemize}
In the next few postures, the position of the arms will change, and the torso posture will be repeated:
\begin{itemize}
\item $\text{ERGD}_{10}$: The subject is standing upright with the forearms bend at 90$^{\circ}$ and the arms raise at the shoulder level, perpendicular to the floor.
\item $\text{ERGD}_{11}$: With the arms at the same position as P10, the subject rotates their torso, and laterally bends to the left.
\item $\text{ERGD}_{12}$: The participant raises their arms perpendicular to the ground while the forearms are fully extended. They proceed by rotating and laterally bending their torso to the left.
\item $\text{ERGD}_{13}$: The subject raises their arms above the head for six seconds.
\item $\text{ERGD}_{14}$: With the arms above the head level, the subject rotates and laterally bends to the left for six seconds.
\end{itemize}
These were all the postures that were assumed from a standing position. The next part describes the postures that will be recorded while the person is seated on a chair.
\begin{itemize}
\item $\text{ERGD}_{15}$: The person is sitting on a chair with the arms relaxed (neutral position).
\item $\text{ERGD}_{16}$: While seated, the subject bends forward at an angle of 60$^{\circ}$ or more.
\item $\text{ERGD}_{17}$: The subject bends forwards at an angle of 60$^{\circ}$ or more while rotating their torso and bending laterally to the left.
\item $\text{ERGD}_{18}$: The subject repeats P17 but has their arms extended in front of them.
\item $\text{ERGD}_{19}$: The subject raises their arms above the head level while they are fully extended.
\item $\text{ERGD}_{20}$: With the arms above the head level, the participant will rotate and laterally bend their torso to the left.
\end{itemize}
Finally, the remaining postures will be performed while the subject is kneeling on their right knee. These are the most ergonomically uncomfortable postures. Beyond that, the upper body options will be the same as before:
\begin{itemize}
\item $\text{ERGD}_{21}$: The subject stays upright.
\item $\text{ERGD}_{22}$: The subject rotates their torso to the left as far as they can, they remain in that position for six seconds.
\item $\text{ERGD}_{23}$: The subject laterally bends their torso to the left.
\item $\text{ERGD}_{24}$: The subject bends forward at an angle larger than 60$^{\circ}$. 
\item $\text{ERGD}_{25}$: While bending the torso at an angle larger than 60$^{\circ}$, the participant rotates and laterally bends their torso to the left.
\item $\text{ERGD}_{26}$: The P25 posture is repeated, but this time, the person’s arms are extended as if to pick something up from the ground.
\item $\text{ERGD}_{27}$: The subject raises their arms to be perpendicular to the ground.
\item $\text{ERGD}_{28}$: With the arms raised, the subject rotates and laterally bends their torso to the left.
\end{itemize}
After completing the recordings, ERGD has examples from the most comfortable positions to some of the most ergonomically improper according to the risk factors defined by EAWS. Though those postures are not in the context of any specific goal, they can act as a baseline to test different methods of an ergonomic assessment. 

\subsubsection{Traditional crafts tasks}
Master artisans and mastic farmers were captured doing their professional tasks in their real workplaces. An additional MoCap session was conducted to capture the simulation of the mastic cultivation task without using any material or tools.

\begin{figure*}[] 
\centering
\subfloat[$\text{SLW}_1$]{\includegraphics[width=0.2\textwidth, height=30mm]{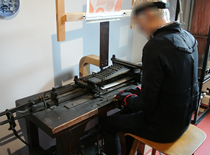}}
\hfil
\subfloat[$\text{SLW}_{3}$]{\includegraphics[width=0.2\textwidth, height=30mm]{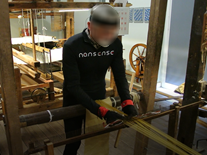}}
\hfil
\subfloat[$\text{SLW}_{4,3}$]{\includegraphics[width=0.2\textwidth, height=30mm]{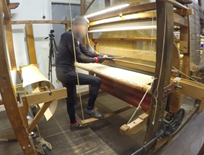}\label{fig:GV4c}}
\hfil
\subfloat[$\text{SLW}_{4,1}$]{\includegraphics[width=0.2\textwidth, height=30mm]{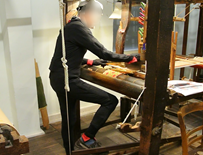}\label{fig:GV4d}}
\caption{Examples of the jacquard weaving gestures recorded.}
\label{fig:GV4}
\end{figure*}

\paragraph{Silk weaving}
In a jacquard loom workshop in Krefeld, Germany, the gestures of a skilled silk weaver were captured. This set of gestures recorded is referenced as \textbf{SLW}, and some examples of these are illustrated in Fig. \ref{fig:GV4}. 
Throughout three days, the expert was recorded performing the following silk weaving-related tasks:
\begin{enumerate}
\item $\text{SLW}_1$: The creation of the punch cards.
\item $\text{SLW}_2$: Wrapping of the beam.
\item $\text{SLW}_3$: Preparation of the beam.
\item $\text{SLW}_{4,1:3}$: Jacquard weaving with looms of different sizes (small, medium, and large).
\end{enumerate}
On the first day, the silk weaver was recorded performing $\text{SLW}_1$, $\text{SLW}_2$, and $\text{SLW}_3$ continuously. The creation of the punch cards was recorded for one hour. Due to the complexity and length of the tasks, the wrapping and preparation of the silk beams were recorded only once, taking about four hours to record. The next two days consisted of continuous recordings of the expert weaving using looms of three different sizes. The recording only stopped when the weaver switched to a different loom. The task of waiving with a loom can be divided into three main gestures ($\text{SLW}_{4,1}$,$\text{SLW}_{4,2}$, and $\text{SLW}_{4,3}$). Firstly, the expert pushes the pedal down with his right leg at the same time that he pushes away the threads with his left hand (the initial posture of the weaver is shown in figures \ref{fig:GV4c} an \ref{fig:GV4d}). Then, by controlling the shuttle that passes the thread horizontally with the right hand, he sends the shuttle to the other side with a quick pulling gesture. Finally, he pulls back the threads with the left hand while simultaneously releasing the pedal with the right leg. This process is repeated up to the end of the piece.

\begin{figure*}[] 
\centering
\subfloat[$\text{GLB}_3$]{\includegraphics[width=0.2\textwidth, height=30mm]{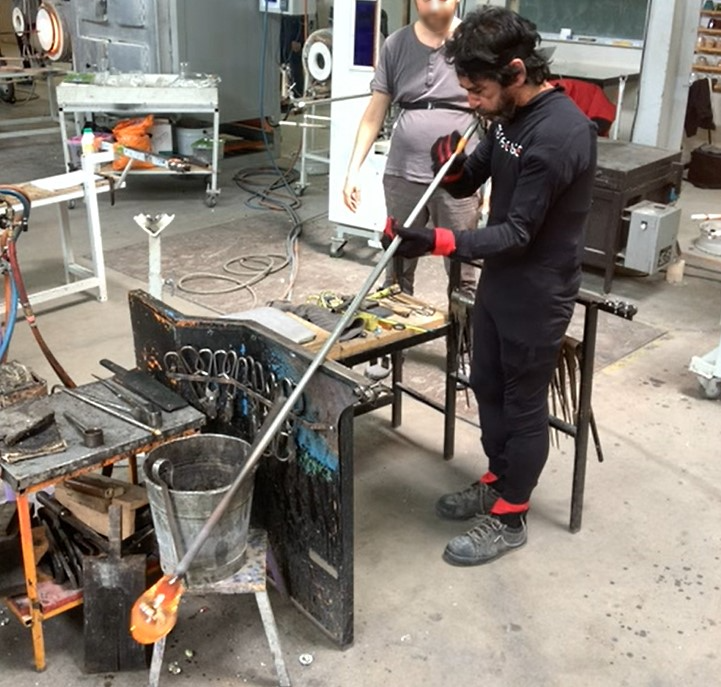}\label{fig:GV5a}}
\hfil
\subfloat[$\text{GLB}_{4}$]{\includegraphics[width=0.2\textwidth, height=30mm]{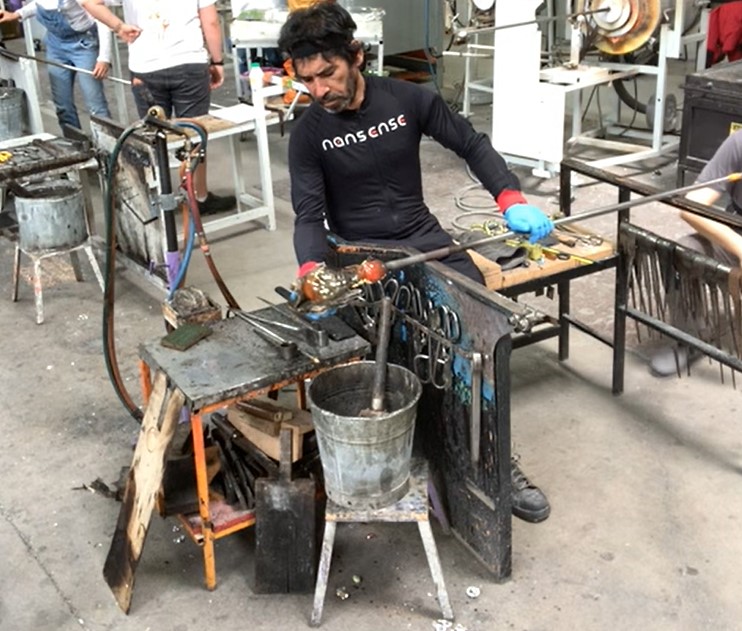}\label{fig:GV5b}}
\hfil
\subfloat[$\text{GLB}_{8}$]{\includegraphics[width=0.2\textwidth, height=30mm]{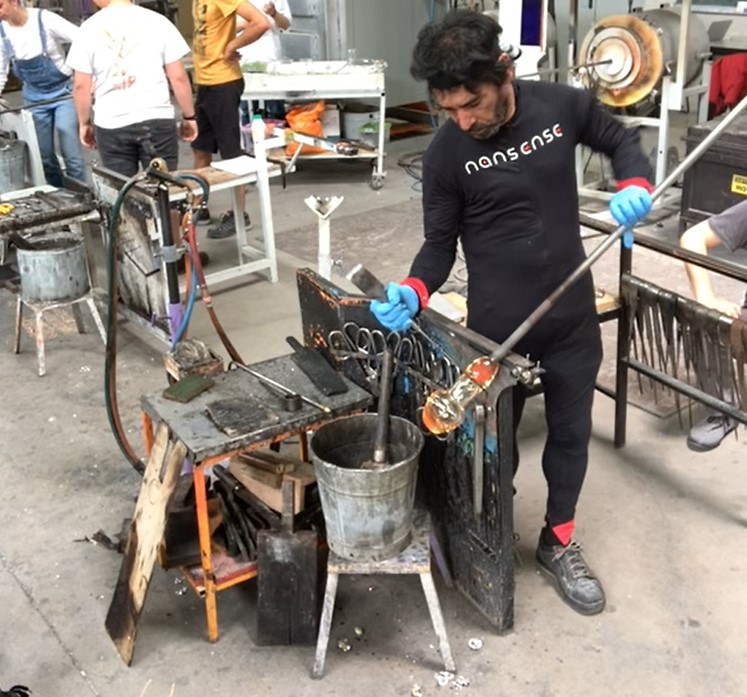}\label{fig:GV5c}}
\hfil
\subfloat[$\text{GLB}_{9}$]{\includegraphics[width=0.2\textwidth, height=30mm]{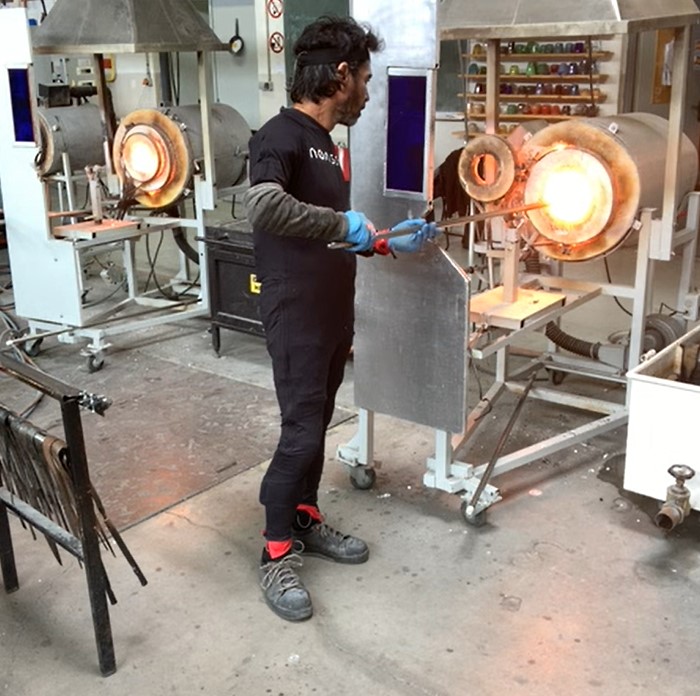}\label{fig:GV5d}}
\caption{Example of gestures captured in a glassblowing workshop.}
\label{fig:GV5}
\end{figure*}

\paragraph{Glassblowing}
The creation of a glass decanter was recorded four times at Vannes-le-Châtel, France, in a European center for research and training in glasswork. Because the temperature of the glass had to be maintained throughout the process, each trial was recorded without pausing between gestures. This resulted in one motion file for each attempt, which starts with collecting the molten glass and finishes when the decanter is left to cool down. The set of gestures composing the process of creating one decanter is denoted as \textbf{GLB}. Fig. \ref{fig:GV5} shows some of the gestures that were recorded during the decanter's fabrication.
The glass decanter was created in three stages. To begin, inflate and shape the molten glass inside the decanter's main body (container). The base was created next, followed by the handle. Next, the expert rolled and shaped the decanter throughout the task to prevent the glass from deforming due to gravity. Finally, an assistant was necessary to blow into the glass while the expert shaped the decanter's main body. 

For shaping the molten glass, the glassblower constantly rotated with his left hand the blowpipe while shaping the glass with his right hand. He utilized various tools with his right hand, including a block (Fig. \ref{fig:GV5b}), jacks (Fig. \ref{fig:GV5c}), soffietta, shears, and metal pencils. These were employed to give the glass the form of the decanter and to add further decorative details. The block is used to maintain the glass's round shape. The jacks are used to shape the decanter's cervix. The shears were utilized to cut the glass and form the decanter's peak. The soffietta forms the decanter's top. Metal pencils were then used to add the handle and extra glass details (cord around the neck) and make the foot (base) of the decanter. Manipulating the tools required constant movement of the right shoulder, right arm, and right forearm. At the same time, the glassblower was seated, rotating back and forth with the left hand the blowpipe on a metal structure. Moving the blowpipe on the metal structure required a small bending to keep the grip of the blowpipe. Placing the handle or shaping the cervix with the jacks required at times for the glassblower to stand up, but he kept moving the blowpipe with the left hand.

While forming the glass, the artisan frequently put the glass on the blowpipe into the furnace (Fig. \ref{fig:GV5d}). He also continuously blew through the blowpipe while holding it horizontally at shoulder height with both arms to maintain the decanter's round shape (Fig. \ref{fig:GV5b}). After finishing, it was passed to a punty to cool down.

\begin{figure*}[]
\centering
\subfloat[$\text{MSC}_3$]{\includegraphics[width=0.2\textwidth, height=30mm]{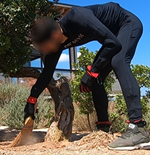}\label{fig:GV6a}}
\hfil
\subfloat[$\text{MSC}_{4}$]{\includegraphics[width=0.2\textwidth, height=30mm]{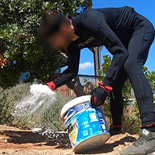}\label{fig:GV6b}}
\hfil
\subfloat[$\text{MSC}_{5}$]{\includegraphics[width=0.2\textwidth, height=30mm]{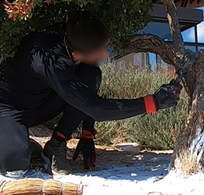}\label{fig:GV6c}}
\hfil
\subfloat[$\text{MSC}_{8}$]{\includegraphics[width=0.2\textwidth, height=30mm]{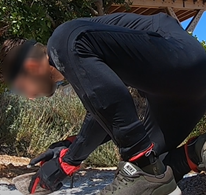}\label{fig:GV6d}}
\caption{Example of gestures captured in the cultivation of mastic.}
\label{fig:GV6}
\end{figure*}

\paragraph{Mastic cultivation}
The cultivation of mastic was recorded in the span of three days in Chios, Greece. The first and second days' recordings were made outside, in front of a mastic tree. The recordings of the last day were simulated inside a room. Each task was divided into separate recordings due to the nature of the cultivation process. This resulted in separate MoCap files for each part of the process. In general, the cultivation of mastic was recorded realistically. However, specific tasks are, in reality, done days or weeks apart or take hours to be completed. As such, the expert was required to demonstrate the gestures briefly while remaining realistic. The gestures recorded from this cultivation process are denoted as \textbf{MSC}. Some gestures that were captured from the mastic farmer are shown in Fig. \ref{fig:GV6}.
The process begins with the preparation of the soil beneath the trees. So that dripping mastic can be easily collected, the earth surrounding the tree is cleaned and the terrain around the tree trunk is leveled. The farmer was recorded using two distinct tools to scrape the soil. The first is an antique agricultural tool (Amia) with a metal head and wooden handle, similar to a trowel. With this one, the farmer scraped the soil on his knees, holding the tool with his right hand. The second tool is a shovel, which allows the farmer to scrape the soil while standing. The farmer then swept the ground with a short broom (Fig. \ref{fig:GV6a}). After preparing the soil, the farmer evenly distributed calcium carbonate ($CaCO_3$) on the ground to create a flat surface. For this task, the farmer knelt and spread the white dust with his right hand while holding the container with his left (Fig. \ref{fig:GV6b}).

The tree is then cut in order to obtain mastic. There are three different tools to do incisions in the tree. The first is a small tool with sharp points at the ends (Kenditiri), the second is another small tool called Timitiri, and the third is a small axe. The farmer was standing while using each tool, but he had to lean over to make the incisions in the tree. The tools were held with the right hand. The next step recorded was the gathering and harvesting of the mastic that had emerged from the tree's wounds. The farmer picked the fallen mastic using a small basket and tweezers (Fig. \ref{fig:GV6d}), and then harvested more resin off the tree with a razor (Fig. \ref{fig:GV6c}). Both gestures required the farmer to bend and manipulate the tool with his right hand.

The farmer wiped the soil to collect it on a metal mesh with a brush. In order to remove dust from the mastic, the mesh is continuously moved (or shifted). The use of two types of mesh was recorded. For all variants, the farmer knelt and moved the mesh with both hands. Finally, a third method for removing the dust from the mastic was recorded: throwing the mastic and dust while standing into the wind.

\section{Data processing and segmentation}\label{app_datprocs}
The processing of the MoCap consisted of two steps. To begin, a low pass filter was applied, followed by the correction of incorrect postures caused by electromagnetic interference or sensors drifting when the recording lasted too long, and calibration was required. A low-pass Butterworth filter was applied to the raw MoCap data to eliminate high-frequency noise. To avoid over-smoothing the data, the cut-off frequency was selected using the power spectrum density of the signal. 

\begin{figure*}[!htbp] 
    \centering
    \includegraphics[width=0.9\textwidth]{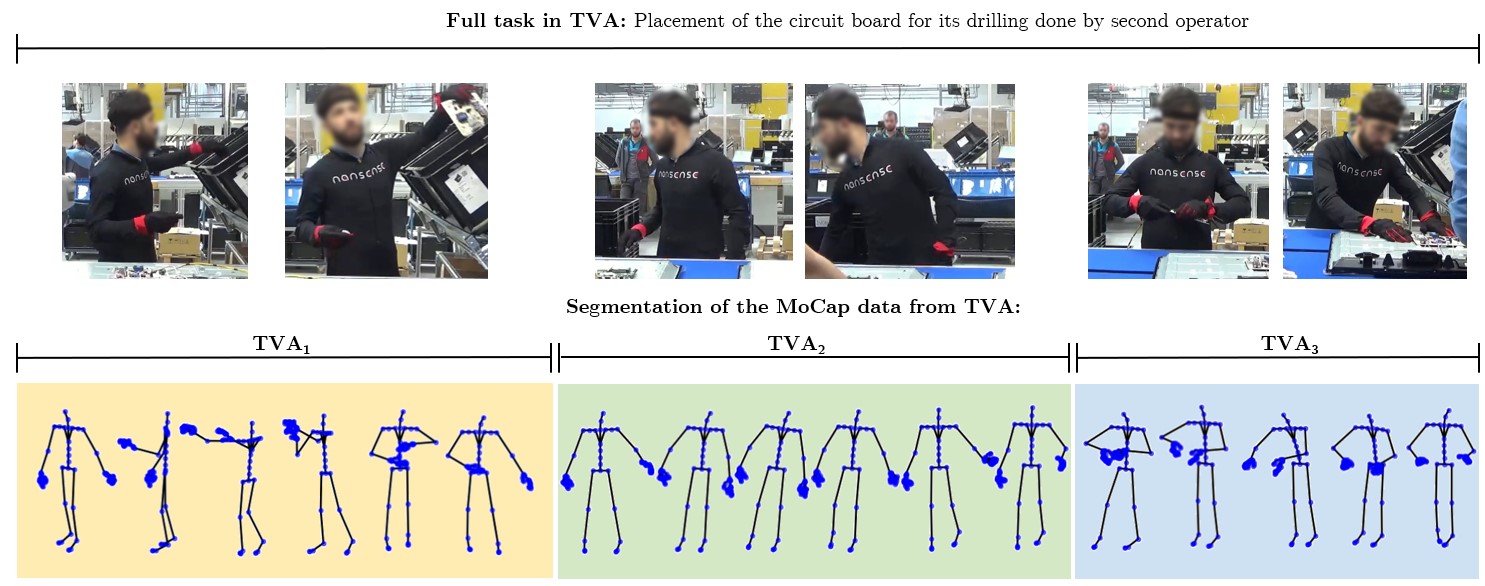} 
    \caption{Gesture segmentation of one repetition of the task of television assembly.}\label{fig:figSegnm}
\end{figure*}

The MoCap system's sensors may drift or be influenced by magnetic disturbances from surrounding metallic objects during the recording process. As a result, occasionally erroneous joint angles were recorded during otherwise precise motion capture. The recordings were adjusted to correct this error using a 3D character animation software\footnote{MotionBuilder, Autodesk Inc., San Rafael, CA. USA}. The software was used to adjust the unrealistic movements based on common sense and video feedback. After adjusting and removing noise from the MoCap data, it was segmented by gestures. Firstly, recordings were collected per task, with one recording representing a whole task; however, these recordings were later segmented by gestures. Fig. \ref{fig:figSegnm} illustrates an example of how the task of television assembly is segmented, extracting the gestures $\text{TVA}_1$, $\text{TVA}_2$, and $\text{TVA}_3$. All the tasks' repetitions in the seven datasets were segmented by gestures or postures for ERGD. A task may contain a single gesture that is performed numerous times, or it may contain additional gestures that are repeated throughout the task.  

The segmentation of the television assembly and packaging is based on repetitions of the gestures given in Section \ref{secP2}. The repetitions segmented from the recordings are shown in Table \ref{tab:tableTV}. For the riveting task, the segmentation of the first gesture consisted of riveting and completing an entire line. The second gesture is to set up the pneumatic hammer for the next line of rivets. Lastly, the final gesture involved placing a bucking bar for an entire line of rivets. Table \ref{tab:TabRivet} illustrates the final segmentation. The recordings of postures with different ergonomic risk levels were segmented into repetitions. Given that ten subjects were recorded assuming 28 poses three times, segmentation produced 840 files containing one repetition of each pose.

The tasks recorded from traditional crafts were segmented by single gestures (as there were repetitions). The resulting segmentation is displayed in tables \ref{tab:tabsilk}, \ref{tab:tabglass}, and \ref{tab:tabmastic}.

\begin{table}[b] 
\centering
\caption{Segmentation of the television assembly task.}
\label{tab:tableTV}
\begin{tabular}{lll} 
\hline
\multicolumn{1}{c}{\textbf{Task}}    & \multicolumn{1}{c}{\textbf{Gesture}} & \multicolumn{1}{c}{\textbf{Repetitions~~}}  \\ 
\hline\hline
\multirow{4}{*}{Television Assembly} & $\text{TVA}_1$~                     & 107                                         \\
                                     & $\text{TVA}_2$                      & 107                                         \\
                                     & $\text{TVA}_3$                      & 108                                         \\
                                     & $\text{TVA}_4$                      & 157                                         \\ 
\hline
\multirow{9}{*}{Packaging}           & $\text{TVP}_1$                      & 8                                           \\
                                     & $\text{TVP}_2$                      & 2                                           \\
                                     & $\text{TVP}_3$                      & 7                                           \\
                                     & $\text{TVP}_4$                      & 5                                           \\
                                     & $\text{TVP}_5$                      & 12                                          \\
                                     & $\text{TVP}_6$                      & 7                                           \\
                                     & $\text{TVP}_7$                      & 7                                           \\
                                     & $\text{TVP}_8$                      & 4                                           \\
                                     & $\text{TVP}_9$                      & 2                                           \\
\hline\hline
\end{tabular}
\end{table}

\begin{table}[b] 
\centering
\caption{Segmentation of the riveting task.}
\label{tab:TabRivet}
\begin{tabular}{lll} 
\hline
\multicolumn{1}{c}{\textbf{Task}} & \multicolumn{1}{c}{\textbf{Gesture}} & \multicolumn{1}{c}{\textbf{Repetitions~~}}  \\ 
\hline\hline
\multirow{3}{*}{Riveting}         & $\text{APA}_1$~                     & 6                                           \\
                                  & $\text{APA}_2$                      & 5                                           \\
                                  & $\text{APA}_3$                      & 8                                           \\
\hline\hline
\end{tabular}
\end{table}

\begin{table}[!htbp] 
\centering
\caption{Segmentation of the silk weaving tasks.}
\label{tab:tabsilk}
\begin{tabular}{lll} 
\hline
\multicolumn{1}{c}{\textbf{Task}}                                                        & \multicolumn{1}{c}{\textbf{Gesture}} & \multicolumn{1}{c}{\textbf{Repetitions}}  \\ 
\hline\hline
Creating a card                                                                          & $\text{SLW}_1$                      & 110                                       \\ 
\hline
\multirow{5}{*}{Beam preparation}                                                        & $\text{SLW}_2,1$                    & 3                                         \\
                                                                                         & $\text{SLW}_2,2$                    & 2                                         \\
                                                                                         & $\text{SLW}_2,3$                    & 4                                         \\
                                                                                         & $\text{SLW}_2,4$                    & 1                                         \\
                                                                                         & $\text{SLW}_2,5$                    & 1                                         \\ 
\hline
Wrapping the beam                                                                        & $\text{SLW}_3$                      & 2                                         \\ 
\hline
\multirow{3}{*}{\begin{tabular}[c]{@{}l@{}}Weaving with \\small size loom\end{tabular}}  & $\text{SLW}_4,1,1$                  & 11                                        \\
                                                                                         & $\text{SLW}_4,1,2$                  & 11                                        \\
                                                                                         & $\text{SLW}_4,1,3$                  & 11                                        \\ 
\hline
\multirow{3}{*}{\begin{tabular}[c]{@{}l@{}}Weaving with \\medium size loom\end{tabular}} & $\text{SLW}_4,2,1$                  & 35                                        \\
                                                                                         & $\text{SLW}_4,2,2$                  & 35                                        \\
                                                                                         & $\text{SLW}_4,2,3$                  & 35                                        \\ 
\hline
\multirow{3}{*}{\begin{tabular}[c]{@{}l@{}}Weaving with \\large size loom\end{tabular}}  & $\text{SLW}_4,3,1$                  & 16                                        \\
                                                                                         & $\text{SLW}_4,3,2$                  & 16                                        \\
                                                                                         & $\text{SLW}_4,3,3$                  & 15                                        \\
\hline\hline
\end{tabular}
\end{table}

\begin{table}[!htbp] 
\centering
\caption{Segmentation of the glassblowing task.}
\label{tab:tabglass}
\begin{tabular}{lll} 
\hline
\multicolumn{1}{c}{\textbf{Task}}             & \multicolumn{1}{c}{\textbf{Gesture}} & \multicolumn{1}{c}{\textbf{Repetitions}}  \\ 
\hline\hline
\multirow{2}{*}{Beak          cutting}        & $\text{GLB}_1$                      & 11                                        \\
                                              & $\text{GLB}_2$                      & 6                                         \\ 
\hline
\multirow{5}{*}{Blowing          and shaping} & $\text{GLB}_3$                      & 5                                         \\
                                              & $\text{GLB}_4$                      & 8                                         \\
                                              & $\text{GLB}_5$                      & 15                                        \\
                                              & $\text{GLB}_6$                      & 7                                         \\
                                              & $\text{GLB}_7$                      & 35                                        \\ 
\hline
Cervix          refining                      & $\text{GLB}_8$                      & 6                                         \\ 
\hline
\multirow{3}{*}{Cord          laying}         & $\text{GLB}_9$                      & 2                                         \\
                                              & $\text{GLB}_10$                     & 8                                         \\
                                              & $\text{GLB}_11$                     & 4                                         \\ 
\hline
Finish          details                       & $\text{GLB}_12$                     & 5                                         \\ 
\hline
\multirow{3}{*}{Handle          laying}       & $\text{GLB}_13$                     & 4                                         \\
                                              & $\text{GLB}_14$                     & 5                                         \\
                                              & $\text{GLB}_15$                     & 4                                         \\ 
\hline
Transfer          to punty                    & $\text{GLB}_16$                     & 4                                         \\ 
\hline
\multirow{2}{*}{Leg          and foot laying} & $\text{GLB}_17$                     & 6                                         \\
                                              & $\text{GLB}_18$                     & 7                                         \\
\hline\hline
\end{tabular}
\end{table}

\begin{table}[!htbp] 
\centering
\caption{Segmentation of the mastic cultivation task.}
\label{tab:tabmastic}
\begin{tabular}{lll} 
\hline
\multicolumn{1}{c}{\textbf{Task}} & \multicolumn{1}{c}{\textbf{Gesture}} & \multicolumn{1}{c}{\textbf{Repetitions}}  \\ 
\hline\hline
Scrapping (New tool)              & $\text{MSC}_1$                      & 3                                         \\
Scrapping (Old tool)              & $\text{MSC}_2$                      & 9                                         \\
Sweeping                          & $\text{MSC}_3$                      & 9                                         \\
Dusting                           & $\text{MSC}_4$                      & 9                                         \\
Embroidery A                      & $\text{MSC}_5$                      & 9                                         \\
Embroidery B                      & $\text{MSC}_6$                      & 3                                         \\
Embroidery with an axe            & $\text{MSC}_7$                      & 3                                         \\
Gathering                         & $\text{MSC}_8$                      & 8                                         \\
Harvesting                        & $\text{MSC}_9$                      & 7                                         \\
Wiping                            & $\text{MSC}_10$                     & 6                                         \\
Shifting A                        & $\text{MSC}_11$                     & 6                                         \\
Shifting B                        & $\text{MSC}_12$                     & 3                                         \\
Cleaning with the wind            & $\text{MSC}_13$                     & 3                                         \\
\hline\hline
\end{tabular}
\end{table}

\begin{figure*}[!htbp] 
    \centering
    \includegraphics[width=0.7\textwidth]{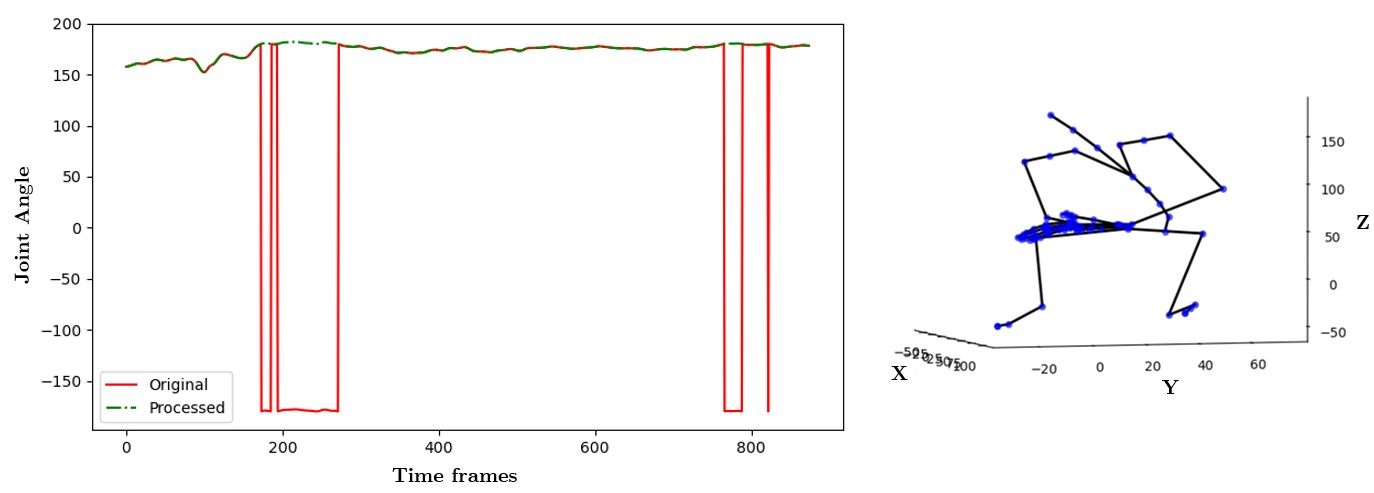}
    \caption{Elimination of data discontinuities for subsequent analysis. The recorded movement corresponds to the Euler angle of the left lower leg on the X-axis (shown in Fig. \ref{fig:sensorsGOM} as $LL$) for the posture $\text{ERGD}_{25}$.}\label{fig:ergd25trans}
\end{figure*}

Only to facilitate the training of the models described in the next sections, the discontinuities of the Euler joint angles present in part of the MoCap files were reduced manually. These discontinuities are dramatic shifts between the values 180$^{\circ}$ and -180$^{\circ}$ in only certain local joint angles. By examining each MoCap file, it was determined to transform the time series with discontinuities to a data of range $[-250^{\circ}, 250^{\circ}]$. Note that this transformation may not be appropriate for new movements recorded with IMUs. Nonetheless, it was sufficient to eliminate most discontinuities in the datasets presented in this paper. Each transformation was documented so that the transformed data may be inversed to Euler angles. An example of these transformations is represented in Fig. \ref{fig:ergd25trans}. The figure illustrates the MoCap data before and after the modifications, as well as the reconstructed skeleton.

The angles from the arms and forearms and one angle of the Hips were mainly the local angles with discontinuities. The angle of the Hips on the Y axis (pointing up, measuring torso rotation) was the most problematic and prone to drifting. The explanation for this could be related to the sensor's position. If the suit is loose, the sensor can produce inaccurate readings. Another factor is that after the suit is turned on and connected to the computer for recording, the subjects must move their entire body to "wake up" the sensors. This sensor was most likely still in an idle state while performing calibrations. Any MoCap file with a distortion caused by drifting or poor calibration was removed from the datasets. The total size of the seven datasets utilized in the following chapters is 5GB. A total of 163,4776 frames, or 5 hours and 2 minutes, make up the segmented gestures with 156 local joint angles measured. 
All BVH files of the seven datasets are accessible in Zenodo\footnote{Benchmark website: https://doi.org/10.5281/zenodo.5356992}.

\section{Analysis of the datasets using analytical models}
Any voluntary movement of the body segments is accomplished via the musculoskeletal system. The musculoskeletal system is an intricate structure comprised of bones, muscles, ligaments, and tendons. Thus, modeling a structure with such complexity is not an easy task. However, even though the musculoskeletal system is primarily responsible for the complexity of human locomotion, it can be acceptable to represent human movements using analytical models that include relevant assumptions about body joint associations and their temporal dependencies. The human movements contained in the seven datasets are then analyzed using analytical models based on the Gesture Operational Model. GOM allows quantifying human dexterity based on the learned parameters of its assumptions. The code for the analysis done in this paper is available in GitHub\footnote{Repository: https://github.com/olivas-bre/GOM.git}

GOM represents human movements using a set of mathematical equations that incorporate assumptions about the stochasticity of human movement and the mediations of body joints. These assumptions allow the proper simulation of human movements using the trained models and explain the evolution of human motion descriptors across time, enabling proactive use of this information. For instance, in human-centered AI technologies, the physical embodiment of humans is the central focus (human-robot collaboration, risk monitoring, or dexterity analysis). Understanding and capturing the dependencies between the movement of different joints is crucial not only for creating more realistic human motion simulations but also for investigating how diverse and intricate full-body human movements are performed. Knowledge of the neurophysiological mechanisms behind complicated dexterity and motor learning may be gleaned from the models. Eventually, the use of such analytic models may enable the development of interdisciplinary frameworks for the research of the process of learning and skill acquisition while performing professional tasks in the industrial or craft sectors. Additionally, they might facilitate research into the key factors that lead to musculoskeletal disorders in ergonomics. 

\subsection{The Gesture Operational Model}
\begin{figure}[b] 
    \centering
    \includegraphics[width=0.8\columnwidth]{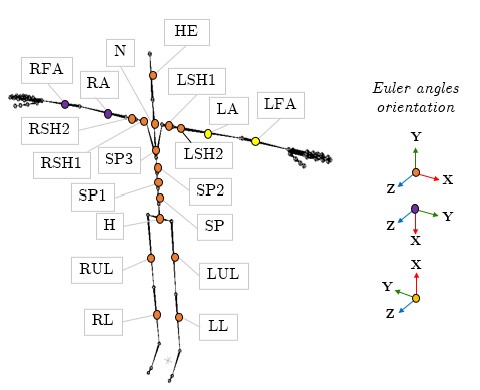} 
    \caption{Location and Euler angle orientation of the sensors that provide the \emph{XYZ} joint angles included in GOM.}\label{fig:sensorsGOM}
\end{figure}

GOM is a mathematical representation of whole-body human movement that takes into account the spatial and temporal dynamics of body joints. The mathematical representation is comprised of a set of models, each of which models a distinct joint motion descriptor using one-shot training with Kalman filters \cite{kalman1960}. The number of models in the equation system of GOM is equal to the number of body joints defined in GOM, multiplied by the number of dimensions the motion descriptor of each joint (e.g., angle or position) is discomposed  (e.g., X, Y, and Z). For this work, the GOM was trained using motion descriptors from only 19 IMUs (out of 52 available in the datasets) for the modeling. Discarding MoCap data from the fingers and feet to simplify the human motion representation. Fig. \ref{fig:sensorsGOM} depicts the sensors' placement, labeling, and orientation. Human postures are expressed as 3D Euler joint angles in order to generate poses with subjects of various morphologies. Unlike joint positions, Euler joint angles are unaffected by identity-specific body shape. Moreover, Euler angles can be intuitively interpreted in the analytical model and provide a more clear illustration of how human movements are conducted.

Thus, 57 models compose the GOMs used in this paper to analyze the full-body movements of every dataset. The 57 models are created through state-space modeling, where endogenous and exogenous data are included in the second-order model of each motion descriptor. For example, while modeling the angle trajectory of the body joint $P_t$ on the \emph{X}-axis ($Px_t$), whose movement is decomposed on \emph{XYZ} axes ($Px_t$, $Py_t$, and $Pz_t$) and has an association with $j$ body parts. The two previous values are integrated into the transition model as shown in Eq. \ref{eq:gomeq1}, where $s_t$ corresponds to the state variable. Then, exogenous data ($u_t$) corresponding to potential intra-joint associations (H2), inter-limb synergies (H3), or intra-limb mediations (H4) are included in the observation model as illustrated in Eq. \ref{eq:gomeq2}.

\begin{equation}\label{eq:gomeq1}
s_t = A s_{t - 1} = \left[ {\begin{array}{*{20}{c}}
{{\alpha _1}}&0\\
0&{{\alpha _2}}
\end{array}} \right]\left[ {\begin{array}{*{20}{c}}
{Px_{1,t-1}}\\
{ - Px_{1,t-2}}
\end{array}} \right]
\end{equation}

\begin{equation}\label{eq:gomeq2}
\begin{aligned}[b]
    Px_{1,t}=\left[ {\begin{array}{*{20}{c}}
1&1
\end{array}} \right]s_t + B u_t = \\
\left[ {\begin{array}{*{20}{c}}
1&1
\end{array}} \right]s_t + {\beta _1}Py_{1,t-1} + {\beta _2}Pz_{1,t-1}+\\
   {\beta_3}Px_{2,t-1}+\cdots+{\beta_n}Px_{j,t-1}
\end{aligned}
\end{equation}
Finally, by merging equations \ref{eq:gomeq1} and \ref{eq:gomeq2}, the state-space representation of the motion descriptor is obtained:

\begin{equation}
\resizebox{0.90\columnwidth}{!}{$
\begin{aligned}[b]
    Px_{1,t}=\underbrace{{\alpha _1}Px_{1,t-1}-{\alpha_2}Px_{1,t-2}}_\text{H1}+ \underbrace{{\beta _1}Py_{1,t-1} + {\beta_2}Pz_{1,t-1}}_\text{H2}+\\
   \underbrace{{\beta_3}Px_{2,t-1}+\cdots+{\beta_n}Px_{j,t-1}}_\text{H3\ or\ H4}
\end{aligned}
$}\label{eq:gomeq3}
\end{equation}
The assumptions and structure of the models are further detailed in \cite{Manitsaris2020}. The constant coefficients $A$ and $B$ of the equation system are estimated using Maximum Likelihood Estimation (MLE) via Kalman filtering. GOMs were trained using a reference gesture of each class, which was determined using the Dynamic Time Warping (DTW) algorithm \cite{Wang2010}. This algorithm measures the similarity between two time series. Then, the gesture repetition closest to all other gesture repetitions of the same class was chosen for one-shot training using an Intel Core i7-8750H CPU. 

Next, Section \ref{secGenHumMocap} discusses the simulation performance of the trained GOMs for every gesture in the seven datasets. Metrics and examples of poses generated are provided in the appendix. These metrics are intended to be used as an initial benchmark of the datasets for comparing the simulation or generation performance of other methods that would use the presented datasets. Later, Section \ref{secDAhum} presents the dexterity analysis of professional gestures using trained GOM representations and how, based on these models, the most significant motion descriptors are identified for modeling and recognizing gestures from a professional task.

\subsection{Generation of full-body movements} \label{secGenHumMocap}
In this section are presented the results of GOM for generating human professional poses. The trained GOM can generate human professional poses by solving its equation system, with each GOM's model predicting one time step per iteration. 

In order to measure the capability of the models in simulating the learned professional gestures, all gesture repetitions were simulated using their respective trained GOM. Then, the Root Mean Squared Error (RMSE) and the Mean Absolute Error (MAE) were calculated for each simulation:
\begin{equation}RMSE=\sqrt{\frac{1}{T}\sum_{t=1}^{T}\left(P_t-{\hat{P}}_t\right)^2}\end{equation}
\begin{equation}MAE=\frac{1}{T}\sum_{t=1}^{T}\left|P_t-{\hat{P}}_t\right|\end{equation}
The real full-body posture corresponds to $P_t$, and ${\hat{P}}_t$ is the simulated movement using the trained GOM. The average of the Theil's inequality coefficients ($U_1$) is also included in the metrics, which coefficient is calculated as follows:
\begin{equation}U_1=\frac{\sqrt{\frac{1}{T}\sum_{t=1}^{T}\left(P_t-{\hat{P}}_t\right)^2}}{\sqrt{\frac{1}{T}\sum_{t=1}^{T}P_t^2}+\sqrt{\frac{1}{T}\sum_{t=1}^{T}{\hat{P}}_t^2}}\end{equation}

\begin{figure*}[!htbp] 
    \centering 
    \includegraphics[width=0.9\textwidth]{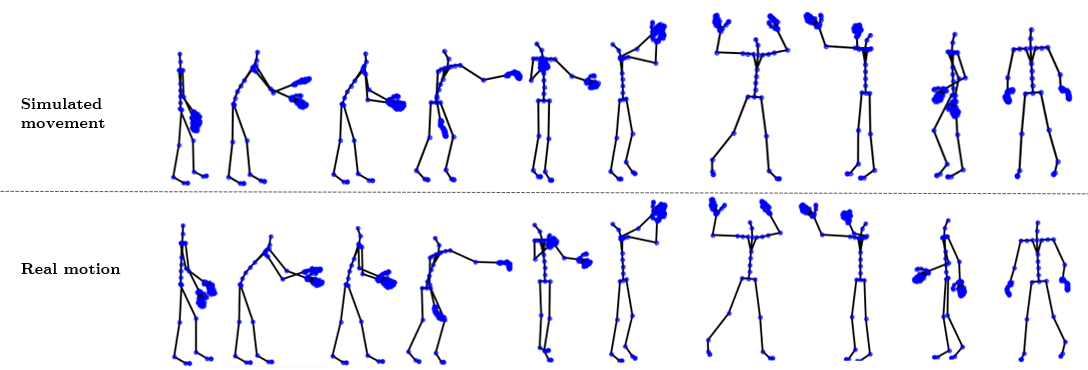} \caption{Visual comparison of generated posture sequences for $\text{TVP}_8$ and its ground-truth. The operator places a television on the third level of a pallet (picture of the recording in Fig. \ref{fig:GV2d}).}\label{fig:postureTVP}
\end{figure*}

\begin{figure*}[!htbp] 
    \centering 
    \includegraphics[width=0.8\textwidth]{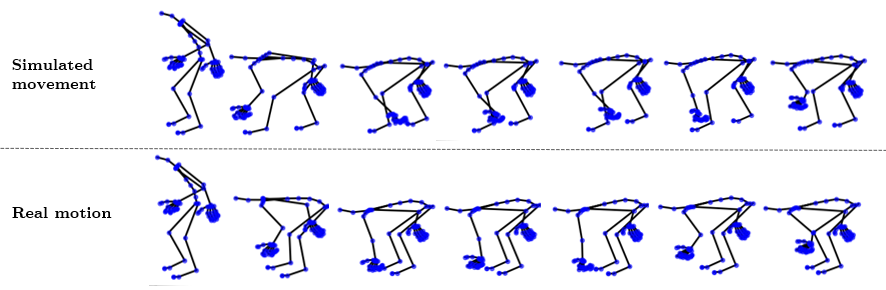} \caption{Visual comparison of generated posture sequences for $\text{MSC}_5$ and its ground-truth. The mastic farmer cuts the root of a mastic tree with a small knife (picture of the recording in Fig. \ref{fig:GV6c}).}\label{fig:postureMSC}
\end{figure*}

For $U_1$, the closer it is to zero, the greater the forecast quality. Tables \ref{tab:resTVA_TVP_APA} to \ref{tab:resMSC} in the appendix show the average measures for each task's gestures. Also, figures \ref{fig:postureGLB} to \ref{fig:postureMSC} illustrate examples of generated postures of a gesture and the real posture sequence.

\subsubsection{Discussion on model simulation}
The results indicate that by solving the simultaneous equations that make up the GOM, it is possible to generate a variety of human postures using Euler joint angles as motion descriptors. GOM is tolerant of minor variations in human movement and offsets between movements of the same class resulting from varying recording conditions (different subjects or different recording days). However, suppose their performance is evaluated regarding their capability to forecast full-body movements accurately. In that case, due to the intra-class variability in some of the professional gestures, there is an increase in the mean of the joint angle errors. The reason is the potential differences between the reference gesture used for the one-shot training of the models and the testing gestures used for the simulation.

For the TVA dataset, the most difficult gestures to simulate accurately were $\text{TVA}_{1}$ and $\text{TVA}_{2}$, which corresponded to gestures in which the operator can move more freely with either the left or right hand to grasp circuit boards or cables. On the other hand, $\text{TVA}_{3}$ and $\text{TVA}_{4}$ correspond to gestures that were easier to replicate for the operator in each iteration, as the circuit board and drilling were performed similarly in each recorded iteration.

GOM provides the best simulation performance for APA, TVP, and ERGD. These gestures and postures had the lowest intra-class variability, given that they were executed in a more controlled environment. In ERGD, for instance, subjects performed various postures in a laboratory while receiving constant instructions on how to execute them. However, as the posture became more complex, such as kneeling with torso and arm movements, the error increased as subjects' movements presented greater variation in how they performed the indicated posture (foot and knee position or arms final position). In the case of APA, the operators were recorded assembling one airplane float, performing the same tasks repeatedly for several hours with only larger variations in $\text{APA}_{2}$ where the operator grabbed the rivets and prepared the pneumatic hammer. In TVP, operators performed the same gestures with high variations only when wrapping the televisions. 

The gestures recorded in industrial settings were easier to simulate since they primarily involved manipulating objects with their hands, in contrast to the gestures performed, for instance, by the craftsmen and farmers, who had to employ their entire bodies to perform their work properly.

The fact that the reference and simulated gestures were executed on different looms may have contributed to the errors in SLW (reference on a large loom and simulated on a medium-size loom). Consequently, pedal height and position variations may have caused larger errors in the movement simulation. Likewise, in the motion simulations of GLB, the skilled glassblower progressively adjusted his posture, even for the same repetitive activity, in order to appropriately shape the molten glass. Consequently, the training gestures for each class of the GLB dataset did not adequately represent all gestures from the same class (high intraclass variance), resulting in a drop in simulation accuracy.

The most challenging gestures to simulate were those involved in mastic cultivation. This may be due to the fact that MSC involves gestures in which the farmer moves while kneeling. In the other six datasets, subjects performed the majority of their tasks while standing. The farmer did not keep the same position of the legs while performing the same gestures; he repositioned the legs while kneeling to improve balance in order to reach the tree or objects.

\subsection{GOM-based dexterity analysis of expert movement} \label{secDAhum}
A statistical analysis is performed on the learned GOM representations to determine the significance of the models' assumptions in relation to the professional gesture. The significant assumptions (motion descriptors) and their learned coefficients are then used to describe the cooperation of the joints to perform the gesture. In addition, by analyzing the p-values of each assumption, the most important motion descriptors for modeling and recognizing human movements from a professional task are found. In many applications of human movement analysis, it is neither feasible nor practical to use full-body MoCap suits. Therefore, to enable the adoption of less intrusive technologies, such as smartphones and smartwatches, a procedure for finding the minimal set of motion descriptors to measure using GOM is also detailed in this paper.
\subsubsection{Statistical analysis and interpretation of the models}
The statistical analysis of three trained motion representations is provided next. To facilitate the visualization of the gesture modeled, a figure with the posture sequence is provided for each example, along with color annotations to highlight the equations' assumptions. GOM's representations are designed to include four assumptions: time-dependent transitions, intra-joint association, inter-limb synergies, and serial and non-serial intra-limb mediations. Each assumption consists of a specific set of parametrized variables (in this case, joint angles) that depict a particular relationship between body joints or a temporal dependency. The notion is to use these parametrized assumptions to describe body dexterity. By examining the computed coefficients and their statistical significance (significant if the p-value is less than 0.05), it can be gleaned how relevant these are according to the gesture modeled and the predicted joint angles. 

\begin{figure*}[!htbp] 
    \centering 
    \includegraphics[width=0.8\textwidth]{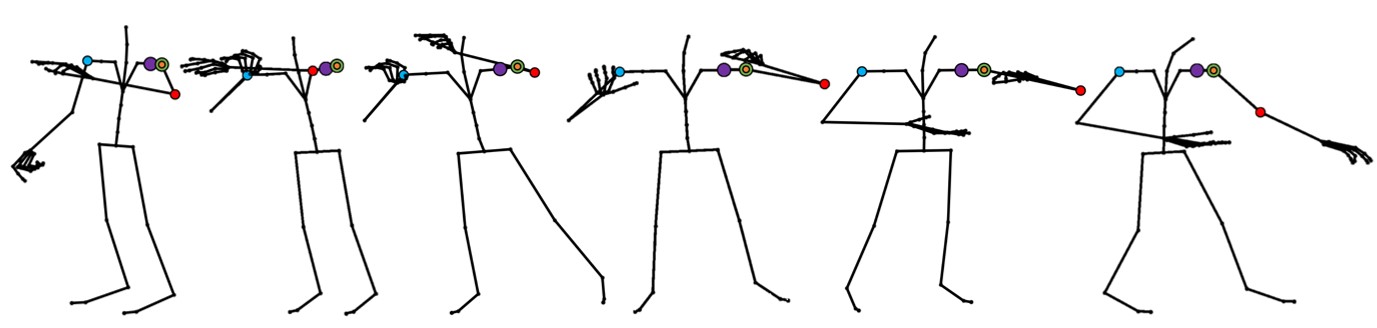} \caption{Illustration of the gesture performed in $\text{TVA}_1$, where the operator grabs from a container a circuit board. The color annotations are based on the assumptions colored in \eqref{eq:expda1}. Colored joints indicate potential dependencies with other motion descriptors incorporated in the model: The orange indicates the transitioning assumption of $RAy$; green reflects the intra-joint association with $RAx$ and $RAz$; blue highlights the inter-limb synergies with $LAy$; purple is the serial intra-limb mediations with $RSH2y$ and red the non-serial intra-limb mediations with $RFAy$. The picture of the recording can also be visualized in Fig. \ref{fig:GV2a}. }\label{fig:timevarModelDA1}
\end{figure*}

The first example illustrates the equation for the joint angle sequence $RAy_{t}$ (right arm on the Y-axis) when performing the gesture $\text{TVA}_1$: (grab a circuit board from a container, shown in Fig. \ref{fig:timevarModelDA1}):
\begin{equation}
\begin{aligned}[b]
    RAy_{t}= \underbrace{(1.010)\textcolor{assumpH1}{RAy_{t-1}}}_\text{p = 0.001}+\underbrace{(-0.076)\textcolor{assumpH1}{RAy_{t-2}}}_\text{p = 0.188}\\
    +\underbrace{(0.720)\textcolor{assumpH2}{RAx_{t-1}}}_\text{p = 0.003} + \underbrace{(1.214)\textcolor{assumpH2}{RAz_{t-1}}}_\text{p < 0.001}\\
    +\underbrace{(-0.324)\textcolor{assumpH3}{LAy_{t-1}}}_\text{p < 0.001}+\underbrace{(6.123)\textcolor{assumpH41}{RSH1y_{t-1}}}_\text{p < 0.001}\\
    +\cdots+\underbrace{(0.555)\textcolor{assumpH42}{RFAy_{t-1}}}_\text{p = 0.009}
\end{aligned}\label{eq:expda1}
\end{equation}

The p-values $< 0.05$ suggest a dependency between the prior value of the dependent variable but not between the value two time steps before. This can imply that the speed of change of the gesture is moderate. If both previous values are significant, this indicates a slow speed movement if neither is a faster one. The movement of the joint RA exhibits an intra-joint association along the X, Y, and Z axes. Inter-limb synergy with $LAy$ (left arm) indicates that $LAy$ follows synergistically $RAy$ when performing the gesture. The movement on $RSH1y$ (right shoulder) and $RFAy$ (right forearm) result in a serial intra-limb mediation. This outcome makes sense, given that most of this arm movement primarily depends on shoulder motions (raising the arm). In addition, if viewing Fig. \ref{fig:GV2a}, the operator must lift the shoulder and bend the forearm to reach the circuit board from the container. The bending of the forearm may explain the statistical significance of $RFAy$.

\begin{figure*}[!htbp] 
    \centering 
    \includegraphics[width=0.9\textwidth]{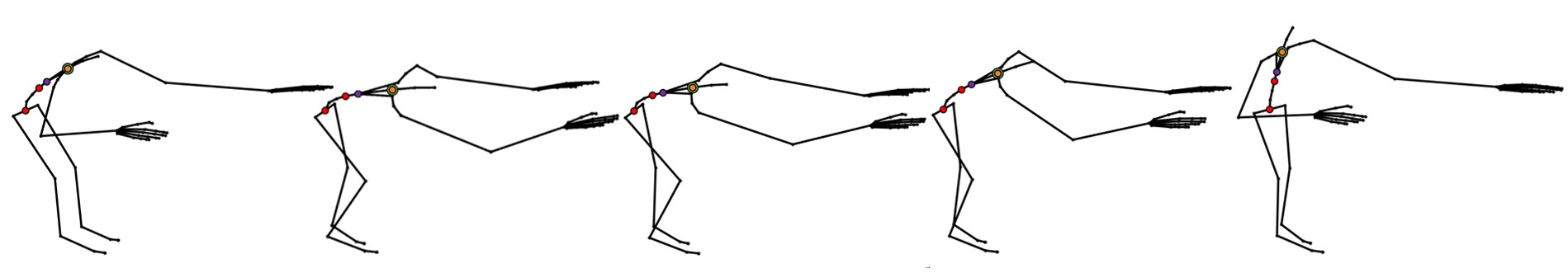} \caption{Illustration of the gesture performed in $\text{APA}_3$, where the operator places the bucking bar to counteract the incoming rivet. The color annotations are based on the assumptions colored in \eqref{expda2}. Colored joints indicate potential dependencies with other motion descriptors incorporated in the model: The orange indicates the transitioning assumption of $Nx$; green reflects the intra-joint association with $Ny$ and $Nz$; purple is the serial intra-limb mediations with $SP3x$ and red the non-serial intra-limb mediations with $SP2x$ and $Hx$. The picture of the recording can also be visualized in Fig. \ref{fig:GV3c}.}\label{fig:timevarModelDA2}
\end{figure*}

The second example is the equation for the joint angle of the neck on the X-axis ($Nx_t$) while performing $\text{APA}_3$ (hold the bucking bar, shown in Fig. \ref{fig:timevarModelDA2}):
\begin{equation}
\begin{aligned}[b]
    Nx_{t} = \underbrace{\left ( 1.020 \right )\textcolor{assumpH1}{Nx_{t-1}}}_\text{p < 0.001}+ \underbrace{\left ( 0.355 \right )\textcolor{assumpH1}{Nx_{t-2}}}_\text{p < 0.024}+ \\ \underbrace{\left ( -1.220 \right )\textcolor{assumpH2}{Ny_{t-1}}}_\text{p < 0.001} 
    + \underbrace{\left ( -0.470 \right )\textcolor{assumpH2}{Nz_{t-1}}}_\text{p < 0.001} \\ 
    +\underbrace{\left ( -0.018 \right )\textcolor{assumpH41}{SP3x_{t-1}}}_\text{p < 0.001} + \underbrace{\left ( -0.010 \right )\textcolor{assumpH42}{SP2x_{t-1}}}_\text{p = 0.002} \\
    +\cdots+ \underbrace{\left ( 0.010 \right )\textcolor{assumpH42}{Hx_{t-1}}}_\text{p = 0.84} 
\end{aligned}\label{expda2}
\end{equation} 
An intra-joint association with $Ny$ and $Nz$ is revealed in \eqref{expda2}, as well as a serial intra-limb mediation with SP3 (upper spine). SP2 (middle spine) exhibits non-serial intra-limb mediation, but H (hips) does not. Holding a bucking bar to counteract a rivet requires bending forward and slightly twisting the upper torso (as illustrated in figures \ref{fig:timevarModelDA2} and \ref{fig:GV3c}), moving along the X-axis and Y-axis of the spine. This movement is reflected in \eqref{expda2}, as the joint angles from SP2 and SP3 on the X and Y axes are statistically significant and relevant to the motion of $Nx$. However, the lack of mediation with H can indicate that the operator tries to maintain his hips static, most likely to keep balance while bending. In addition, the subject had to rotate the neck to see where to position the bucking bar; thus, this is consistent with the intra-joint association indicated by the p-values of $Ny$ and $Nz$. At last, the gesture is performed at a low pace as both transition assumptions are significant.

\begin{figure*}[]
    \centering 
    \includegraphics[width=0.9\textwidth]{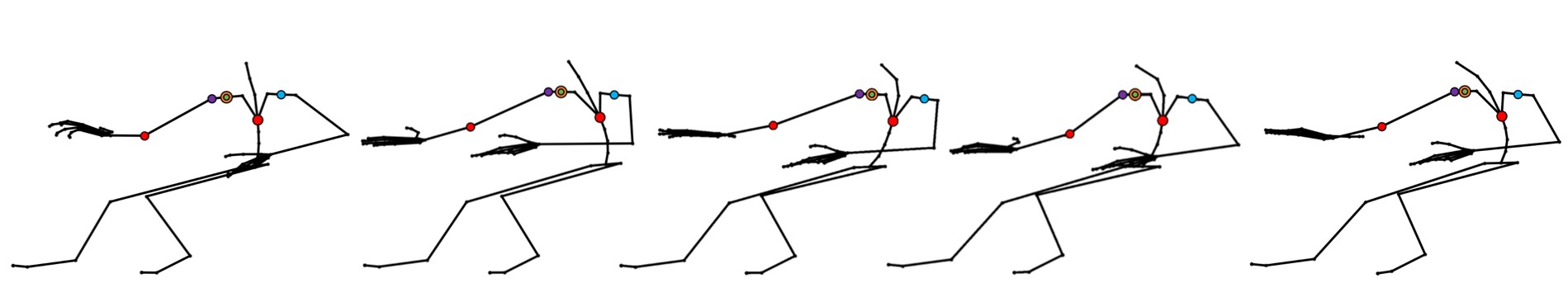} \caption{Illustration of the gesture performed in $\text{GLB}_4$, where the expert glassblower shapes the decanter curve with a block and simultaneously rotates the blowpipe back and forward. The color annotations are based on the assumptions colored in \eqref{eq:expda3}. Colored joints indicate potential dependencies with other motion descriptors incorporated in the model: The orange indicates the transitioning assumption of $LSH2x$; green reflects the intra-joint association with $LSH2y$ and $LSH2z$; blue highlights the inter-limb synergies with $RSH2x$; purple is the serial intra-limb mediations with $LAx$ and red the non-serial intra-limb mediations with $LFAx$ and $SP3x$. The picture of the recording is shown in Fig. \ref{fig:GV5a}.}\label{fig:timevarModelDA3}
\end{figure*}

The last example is an equation learned with the gesture $\text{GLB}_{4}$ (shape the decanter curves with a block, as depicted in Fig. \ref{fig:timevarModelDA3}), and represents the joint angle on the X-axis of the left shoulder ($LSH2x_{t}$). More precisely, this equation simulates the movement of the left clavicle:
\begin{equation}\label{eq:expda3}
\begin{aligned}[b]
    LSH2x_{t}=\underbrace{(1.877)\textcolor{assumpH1}{LSH2x_{t-1}}}_\text{p < 0.001}+\underbrace{(-0.913)\textcolor{assumpH1}{LSH2x_{t-2}}}_\text{p < 0.001}\\
    +\underbrace{(0.292)\textcolor{assumpH2}{LSH2y_{t-1}}}_\text{p = 0.002} + \underbrace{(0.252)\textcolor{assumpH2}{LSH2z_{t-1}}}_\text{p = 0.004}\\
    +\underbrace{(0.145)\textcolor{assumpH3}{RSH2x_{t-1}}}_\text{p = 0.014}+\underbrace{(0.36)\textcolor{assumpH41}{LAx_{t-1}}}_\text{0.004}\\
    +\cdots+\underbrace{(0.016)\textcolor{assumpH42}{LFAx_{t-1}}}_\text{p = 0.030}+\underbrace{(-0.543)\textcolor{assumpH42}{SP3x_{t-1}}}_\text{p = 0.049}
\end{aligned}
\end{equation}
The statistical analysis of \eqref{eq:expda3} reveals a temporal dependence (slow movement); intra-joint association ($LSH2y$ and $LSH2z$); inter-limb synergy with the right shoulder; serial intra-limb mediation with the left arm ($LAx$), and non-serial mediation with the left forearm ($LFAx$). SP3 is considered marginally significant, as this study uses a p-value threshold of 0.05 to determine significance. 

\begin{figure*}[!htbp] 
    \centering 
    \includegraphics[width=\textwidth]{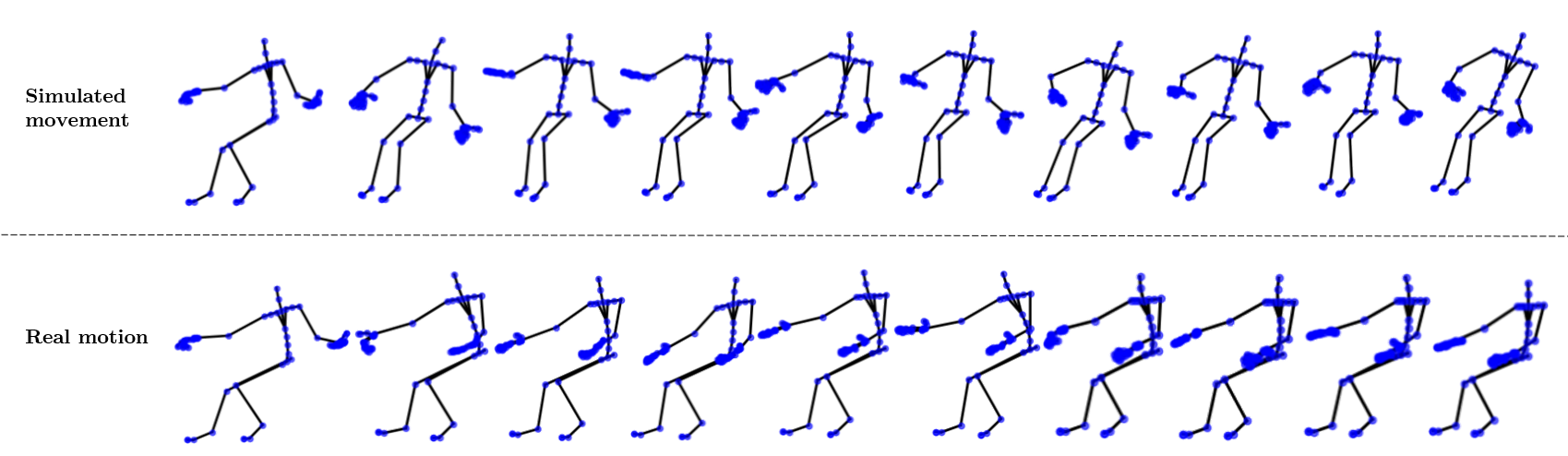} \caption{Visual comparison of generated posture sequences for $\text{GLB}_4$ and its ground-truth. The glassblower rotates the blowpipe with the left hand while shaping the glass with the right (the recording of the glassblower is shown in Fig. \ref{fig:GV5a}).}\label{fig:postureGLB}
\end{figure*}

To shape the decanter correctly, both arms must work together during this gesture. This is evident by the presence of an inter-limb synergy in \eqref{eq:expda3}. Accordingly, the joint angles of the right shoulder contribute to the response of the left shoulder, as the glassblower forms the decanter's curves with the right arm while rolling the blowpipe with the left. Furthermore, the expert mostly maintains the torso straight during this gesture, as seen in figures \ref{fig:postureGLB} and \ref{fig:GV5a}. Yet, when he rotates the blowpipe forward, there is a slight tilt of the torso to maintain grip on the blowpipe; this could indicate a high p-value for SP3, but not as high to not be significant for the left shoulder movement.

As shown in these previous examples, GOM can provide quantitive information that is not directly observable about how the experts perform the modeled gesture and allow interpretation of how the joints collaborate to perform specific joint motion trajectories in order to perform the intended task. Calculating the significance of the assumptions highlighted the joints that are critical in the gesture and their influence on the movement of other joints. This information can later be utilized to test skill acquisition strategies. For example, a novice can learn to make precise gestures by minimizing the variability of their motion representations compared to those of professional artisans or operators. Moreover, for ergonomics, the proposed motion representation would allow analysts to comprehend how the full-body moves when doing ergonomically dangerous movements versus safe movements and to design work environments and tasks that are less likely to result in injury or discomfort. A direction would be to identify which joints have the highest impact while executing risky movements, and that should be monitored to reduce the ergonomic risk of the professional task.

\subsubsection{Selection of most significant motion descriptors per dataset}
For selecting the essential inertial sensors to use for the gesture recognition of each professional task, the number of times a motion descriptor (assumption) is statistically significant for all equations that comprise GOM is counted. Then, different combinations of descriptors considered most frequently significant for measuring the arm, spine, and legs were utilized for training in an all-shots approach. Because a single inertial sensor gives three joint angles, all of the sensor's joint angles were used for recognition if at least one was among the joint angles that were more often significant in all gestures of a dataset.

For the recognition of human gestures utilizing different sensor combinations, Hidden Markov Models (HMM) were trained using a 10-fold cross-validation. In order to properly train the HMM, a gesture vocabulary containing the gestures with the most iterations was specified for each dataset. The total number of gesture classes for TVA, APA, and ERGD were four, three, and 28, respectively. The TVP, GLB, and MSC gesture vocabularies contained only gestures with at least seven repetitions. Therefore, their respective gesture vocabularies included five, seven, and six classes of gestures. Regarding SLW, the gesture vocabulary consisted of only three classes of silk weaving on a loom. Despite the differences in loom size, the gestures used to weave on a small, medium, and large loom are similar. Therefore, they were combined into three classes for the gesture recognition problem.

The ergodic and left-to-right HMM topologies, along with a different number of hidden states, were evaluated to determine the best settings for the gesture vocabulary defined in each dataset. The performance metrics utilized were accuracy and F1-score, the last being the harmonic mean of precision and recall.
\begin{table*} 
\centering
\caption{Recognition performance with each configuration of sensors.}
\label{tab:RP}
\begin{tabular}{lllll} 
\hline
\multicolumn{1}{c}{\textbf{Gesture Vocabularies}} & \multicolumn{1}{c}{\# of classes} & \multicolumn{1}{c}{\textbf{Sensors}} & \textbf{Accuracy (\%)} & \multicolumn{1}{c}{\textbf{F1-Score (\%)}}  \\ 
\hline\hline
\multirow{3}{*}{TVA}                              & \multirow{3}{*}{4}                                                                                            & All 52 sensors                       & \underline{0.967}          & \underline{0.966}                              \\
                                                  &                                                                                                               & Two sensors: RFA, H                          & 0.891                  & 0.871                                      \\
                                                  &                                                                                                               & LA, SP1, RUL                         & 0.910                  & 0.902                                      \\ 
\hline
\multirow{3}{*}{TVP}                              & \multirow{3}{*}{5}                                                                                            & All 52 sensors                       & \underline{0.950}          & \underline{0.916}                              \\
                                                  &                                                                                                               & Two sensors: RFA, H                          & 0.888                  & 0.850                                      \\
                                                  &                                                                                                               & RSH1, LFA, SP2                       & 0.901                  & 0.843                                      \\ 
\hline
\multirow{3}{*}{APA}                              & \multirow{3}{*}{3}                                                                                            & All 52 sensors                       & 0.850                  & 0.833                                      \\
                                                  &                                                                                                               & Two sensors: RFA, H                          & 0.750                  & 0.701                                      \\
                                                  &                                                                                                               & RA, LSH1, LSH2,SP3, SP2,LUL,RUL      & \underline{0.915}          & \underline{0.901}                              \\ 
\hline
\multirow{3}{*}{ERGD}                             & \multirow{3}{*}{28}                                                                                           & All 52 sensors                       & 0.916                  & 0.902                                      \\
                                                  &                                                                                                               & Two sensors: RFA, H                          & 0.738                  & 0.735                                      \\
                                                  &                                                                                                               & LA, RSH1, LFA,SP2,~RUL               & \underline{0.927}          & \underline{0.917}                              \\ 
\hline
\multirow{3}{*}{SLW}                              & \multirow{3}{*}{3}                                                                                            & All 52 sensors                       & \underline{0.954}          & \underline{0.943}                              \\
                                                  &                                                                                                               & Two sensors: RFA, H                          & 0.620                  & 0.610                                      \\
                                                  &                                                                                                               & RSH1, LSH1, HE,LUL, RL               & 0.909                  & 0.892                                      \\ 
\hline
\multirow{3}{*}{GLB}                              & \multirow{3}{*}{7}                                                                                            & All 52 sensors                       & \underline{0.917}          & 0.816                                      \\
                                                  &                                                                                                               & Two sensors: RFA, H                          & 0.810                  & 0.801                                      \\
                                                  &                                                                                                               & LSH2, RFA, H, SP3                    & 0.842                  & \underline{0.850}                              \\ 
\hline
\multirow{3}{*}{MSC}                              & \multirow{3}{*}{6}                                                                                            & All 52 sensors                       & \underline{0.866}          & \underline{0.866}                              \\
                                                  &                                                                                                               & Two sensors: RFA, H                          & 0.799                  & 0.750                                      \\
                                                  &                                                                                                               & LSH1, SP3, LUL, LL                   & \underline{0.866}          & \underline{0.866}                              \\
\hline\hline
\end{tabular}
\end{table*}

Left-to-right HMM topology produced the best results for all recognition problems. Concerning the number of hidden states, it was defined for the HMMs of TVA and ERGD with seven states, TVP with six states, APA, GLB, and MSW with eight states, and SLW with three states. The sensor configurations that obtained the best recognition results with each dataset's gestures are presented in Table \ref{tab:RP}, along with the highest recognition accuracy and F1-score achieved.

In Table \ref{tab:RP}, it can be observed that using MoCap data just from the selected sensors yielded a performance that was comparable to or better than that obtained using all MoCap data from the 52 inertial sensors. In the case of TVA and TVP, just three sensors were selected based on the trained models of GOM, which represents 5.76\% of the MoCap data acquired and causes only a minor loss in accuracy and F1-score. With APA and ERGD, only 13.46\% and 9.61\% of the MoCap data were selected and utilized to achieve a higher recognition performance than the other two configurations of sensors.

The APA gestures were the most difficult to recognize, requiring a greater number of sensors for effective recognition. This could be due to the fact that the gestures in this vocabulary are more complex and prolonged. The most problematic gesture to model and recognize was $\text{APA}_{2}$, which was expected given that its execution varied the most among the three classes (high intra-class variance). The operator did not prepare the material identically for each repetition. In certain repetitions, the operator was slower than usual because he required more time to adjust the pneumatic hammer or to prepare additional rivets. Furthermore, since only one airplane structure was built for this dataset, there is a substantial intra-class variance. There were no repetitions in which the pneumatic hammer was positioned in the same location more than once.

Regarding recognizing gestures from traditional crafts, for GLB and MSC, the selected sensors, consisting of 7.69\% of the MoCap data, yielded comparable results to those obtained with data from all sensors. For SLW, 9.61\% of the MoCap data was selected for the recognition problem, resulting in a performance drop of about 0.05 in both metrics with respect to the configuration with all 52 sensors. The two-sensor configuration's poor performance for SLW could be attributed to its difficulty in distinguishing movements related to the shoulder (throwing of the shuttle) and the leg, as motion data from the hips and left forearm only were insufficient.

\section{Conclusion}
This paper presented seven datasets: TVA, TVP, APA, ERGD, SLW, GLB, and MSC. Most publicly available datasets contain simulated movements performed in a laboratory and related to everyday activities or sports. Therefore, new datasets were created containing gestures performed in professional tasks either from the industry or crafts workshops. These were recorded with actual operators and experts in their real workplace scenarios using an inertial full-body suit of 52 sensors. The aim was to test human motion models with these complex gestures and extract information regarding the dexterity, skill, and know-how related to the adequate use of tangible elements such as materials and tools. Each professional task was segmented by repetitions, and discontinuities were reduced to improve the modeling of the gestures in the analysis done for this benchmark.

The presented human movement analysis comprised the use of GOM to simulate the recorded professional tasks and a body dexterity analysis based on the trained motion representations. The purpose was to employ the trained motion models to observe and quantify the manifestation of skill in industrial operators and expert artisans.  The parameters of the train models provided information about how a person moves in order to achieve a specific goal, such as assembling a TV or making a specific piece of glass. In the future, multidisciplinary frameworks might be built to study how people learn and get better at industrial or craft tasks by looking at the trained analytical models of experts and beginners. Furthermore, GOM could be used to investigate the biomechanical risk factors that lead to work-related musculoskeletal disorders by comparing motion representations from safe and hazardous movements.

Finally, the minimum number of inertial sensors and their location for capturing and accurately recognizing the gestures of each recorded professional task is presented. As stated before, employing a full-body MoCap suit in many human movement analysis applications is neither feasible nor practicable. Determining the minimal motion descriptors to measure allows for the adoption of less invasive technologies, such as smartphones and smartwatches, that could also measure these motion descriptors.

\section*{Acknowledgment}
We would like to thank Jean-Pierre Mateus, the Pireaus foundation, the Haus der Seidenkultur museum, and the Romaero and Arçelik factories for contributing to the creation of the datasets.

\appendices

\section{Forecasting performance measures for each datataset}
\begin{table}[!htbp]  
\centering
\caption{Simulation performance for datasets TVA, TVP, and APA.}\label{tab:resTVA_TVP_APA}
\begin{tabular}{lllll} 
\hline
\multicolumn{1}{c}{\textbf{Dataset}} & \multicolumn{1}{c}{\textbf{Gesture}} & \multicolumn{1}{c}{\textbf{RMSE}} & \multicolumn{1}{c}{\textbf{MAE}} & \multicolumn{1}{c}{\textbf{Avg $U_1$}}  \\ 
\hline\hline
\multirow{4}{*}{TVA}                 & $\text{TVA}_{1}$                                & 52.637                            & 25.610                           & 0.508                                \\
                                     & $\text{TVA}_{2}$                                & 65.893                            & 37.600                           & 0.825                                \\
                                     & $\text{TVA}_{3}$                                & 6.836                             & 2.742                            & 0.116                                \\
                                     & $\text{TVA}_{4}$                                & 4.385                             & 1.368                            & 0.088                                \\ 
\hline
\multirow{9}{*}{TVP}                 & $\text{TVP}_{1}$                                & 10.241                            & 2.050                            & 0.046                                \\
                                     & $\text{TVP}_{2}$                                 & 15.746                            & 4.779                            & 0.078                                \\
                                     & $\text{TVP}_{3}$                                 & 10.164                            & 5.545                            & 0.099                                \\
                                     & $\text{TVP}_{4}$                                 & 9.657                             & 2.267                            & 0.060                                \\
                                     & $\text{TVP}_{5}$                                 & 13.669                            & 12.305                           & 0.232                                \\
                                     & $\text{TVP}_{6}$                                 & 18.192                            & 15.816                           & 0.202                                \\
                                     & $\text{TVP}_{7}$                                 & 21.746                            & 16.452                           & 0.219                                \\
                                     & $\text{TVP}_{8}$                                 & 16.898                            & 4.132                            & 0.082                                \\
                                     & $\text{TVP}_{9}$                                 & 27.492                            & 8.178                            & 0.105                                \\ 
\hline
\multirow{3}{*}{APA}                 & $\text{APA}_{1}$                                & 8.311                             & 1.550                            & 0.110                                \\
                                     & $\text{APA}_{2}$                                & 42.558                            & 7.286                            & 0.480                                \\
                                     & $\text{APA}_{3}$                                & 2.575                             & 1.100                            & 0.042                                \\
\hline\hline
\end{tabular}
\end{table}
\begin{table}[!htbp] 
\centering
\caption{Simulation performance for the dataset ERGD.}
\begin{tabular}{lllll} 
\hline
\multicolumn{1}{c}{\textbf{Dataset}} & \multicolumn{1}{c}{\textbf{Gesture}} & \multicolumn{1}{c}{\textbf{RMSE}} & \multicolumn{1}{c}{\textbf{MAE}} & \multicolumn{1}{c}{\textbf{Avg $U_1$}}  \\ 
\hline\hline
\multirow{28}{*}{ERGD}               & $\text{ERGD}_{1}$                               & 7.347                             & 6.516                            & 0.323                                \\
                                     & $\text{ERGD}_{2}$                               & 9.793                             & 4.378                            & 0.165                                \\
                                     & $\text{ERGD}_{3}$                               & 4.120                             & 2.633                            & 0.136                                \\
                                     & $\text{ERGD}_{4}$                               & 5.031                             & 3.332                            & 0.162                                \\
                                     & $\text{ERGD}_{5}$                               & 5.066                             & 2.759                            & 0.137                                \\
                                     & $\text{ERGD}_{6}$                               & 7.596                             & 4.750                            & 0.137                                \\
                                     & $\text{ERGD}_{7}$                               & 6.159                             & 3.657                            & 0.165                                \\
                                     & $\text{ERGD}_{8}$                               & 4.780                             & 3.090                            & 0.108                                \\
                                     & $\text{ERGD}_{9}$                               & 4.914                             & 3.063                            & 0.092                                \\
                                     & $\text{ERGD}_{10}$                              & 22.163                            & 18.234                           & 0.247                                \\
                                     & $\text{ERGD}_{11}$                              & 28.283                            & 22.209                           & 0.264                                \\
                                     & $\text{ERGD}_{12}$                              & 11.698                            & 6.586                            & 0.177                                \\
                                     & $\text{ERGD}_{13}$                              & 11.316                            & 4.952                            & 0.172                                \\
                                     & $\text{ERGD}_{14}$                              & 35.360                            & 16.796                           & 0.225                                \\
                                     & $\text{ERGD}_{15}$                              & 32.047                            & 17.635                           & 0.224                                \\
                                     & $\text{ERGD}_{16}$                              & 16.538                            & 8.923                            & 0.196                                \\
                                     & $\text{ERGD}_{17}$                              & 18.340                            & 10.190                           & 0.215                                \\
                                     & $\text{ERGD}_{18}$                              & 24.824                            & 18.354                           & 0.211                                \\
                                     & $\text{ERGD}_{19}$                              & 25.478                            & 22.967                           & 0.306                                \\
                                     & $\text{ERGD}_{20}$                              & 34.291                            & 18.294                           & 0.221                                \\
                                     & $\text{ERGD}_{21}$                              & 30.535                            & 15.333                           & 0.307                                \\
                                     & $\text{ERGD}_{22}$                              & 24.023                            & 16.639                           & 0.351                                \\
                                     & $\text{ERGD}_{23}$                              & 22.024                            & 15.077                           & 0.205                                \\
                                     & $\text{ERGD}_{24}$                              & 39.893                            & 23.038                           & 0.348                                \\
                                     & $\text{ERGD}_{25}$                              & 44.179                            & 27.285                           & 0.363                                \\
                                     & $\text{ERGD}_{26}$                              & 44.331                            & 29.911                           & 0.473                                \\
                                     & $\text{ERGD}_{27}$                              & 43.128                            & 23.421                           & 0.401                                \\
                                     & $\text{ERGD}_{28}$                              & 45.235                            & 26.899                           & 0.459                                \\
\hline\hline
\end{tabular}
\end{table}
\begin{table}[!htbp] 
\centering
\caption{Simulation performance for the dataset SLW.}
\begin{tabular}{lllll} 
\hline
\multicolumn{1}{c}{\textbf{Dataset}} & \multicolumn{1}{c}{\textbf{Gesture}} & \multicolumn{1}{c}{\textbf{RMSE}} & \multicolumn{1}{c}{\textbf{MAE}} & \multicolumn{1}{c}{\textbf{Avg $U_1$}}  \\ 
\hline\hline
\multirow{16}{*}{SLW}                & $\text{SLW}_{1}$                                & 16.518                            & 8.841                            & 0.063                                \\
                                     & $\text{SLW}_{2,1}$                              & 14.563                            & 7.411                            & 0.079                                \\
                                     & $\text{SLW}_{2,2}$                              & 14.441                            & 7.104                            & 0.008                                \\
                                     & $\text{SLW}_{2,3}$                              & 20.840                            & 10.217                           & 0.016                                \\
                                     & $\text{SLW}_{2,4}$                              & 16.551                            & 3.097                            & 0.005                                \\
                                     & $\text{SLW}_{2,5}$                              & 6.805                             & 3.120                            & 0.024                                \\
                                     & $\text{SLW}_{3}$                                & 21.907                            & 10.666                           & 0.082                                \\
                                     & $\text{SLW}_{4,1,1}$                            & 28.765                            & 15.427                           & 0.393                                \\
                                     & $\text{SLW}_{4,1,2}$                            & 27.136                            & 13.489                           & 0.180                                \\
                                     & $\text{SLW}_{4,1,3}$                            & 52.251                            & 30.356                           & 0.532                                \\
                                     & $\text{SLW}_{4,2,1}$                            & 31.695                            & 15.635                           & 0.234                                \\
                                     & $\text{SLW}_{4,2,2}$                            & 50.182                            & 25.050                           & 0.318                                \\
                                     & $\text{SLW}_{4,2,3}$                            & 47.670                            & 18.471                           & 0.337                                \\
                                     & $\text{SLW}_{4,3,1}$                            & 30.179                            & 14.747                           & 0.265                                \\
                                     & $\text{SLW}_{4,3,2}$                            & 38.361                            & 18.740                           & 0.293                                \\
                                     & $\text{SLW}_{4,3,3}$                            & 42.731                            & 25.895                           & 0.383                                \\
\hline\hline
\end{tabular}
\end{table}
\begin{table}[!htbp] 
\centering
\caption{Simulation performance for the dataset GLB.}
\begin{tabular}{lllll} 
\hline
\multicolumn{1}{c}{\textbf{Dataset}} & \multicolumn{1}{c}{\textbf{Gesture}} & \multicolumn{1}{c}{\textbf{RMSE}} & \multicolumn{1}{c}{\textbf{MAE}} & \multicolumn{1}{c}{\textbf{Avg $U_1$}}  \\ 
\hline\hline
\multirow{18}{*}{GLB}                & $\text{GLB}_{1}$                                & 68.644                            & 36.340                           & 0.421                                \\
                                     & $\text{GLB}_{2}$                                & 7.921                             & 2.929                            & 0.125                                \\
                                     & $\text{GLB}_{3}$                                & 43.928                            & 21.410                           & 0.294                                \\
                                     & $\text{GLB}_{4}$                                & 23.980                            & 11.502                           & 0.156                                \\
                                     & $\text{GLB}_{5}$                                & 49.547                            & 24.251                           & 0.314                                \\
                                     & $\text{GLB}_{6}$                                & 41.230                            & 17.904                           & 0.351                                \\
                                     & $\text{GLB}_{7}$                                & 10.292                            & 5.213                            & 0.146                                \\
                                     & $\text{GLB}_{8}$                                & 65.316                            & 28.045                           & 0.394                                \\
                                     & $\text{GLB}_{9}$                                & 62.949                            & 21.055                           & 0.297                                \\
                                     & $\text{GLB}_{10}$                               & 59.623                            & 25.070                           & 0.290                                \\
                                     & $\text{GLB}_{11}$                               & 51.223                            & 32.240                           & 0.534                                \\
                                     & $\text{GLB}_{12}$                               & 16.418                            & 7.859                            & 0.176                                \\
                                     & $\text{GLB}_{13}$                               & 38.091                            & 18.714                           & 0.160                                \\
                                     & $\text{GLB}_{14}$                               & 88.048                            & 33.678                           & 0.519                                \\
                                     & $\text{GLB}_{15}$                               & 20.144                            & 9.203                            & 0.198                                \\
                                     & $\text{GLB}_{16}$                               & 11.893                            & 7.300                            & 0.176                                \\
                                     & $\text{GLB}_{17}$                               & 46.920                            & 15.606                           & 0.230                                \\
                                     & $\text{GLB}_{18}$                               & 66.354                            & 27.487                           & 0.475                                \\
\hline\hline
\end{tabular}
\end{table}
\begin{table}[!htbp] 
\centering
\caption{Simulation performance for the dataset MSC.}\label{tab:resMSC}
\begin{tabular}{lllll} 
\hline
\multicolumn{1}{c}{\textbf{Dataset}} & \multicolumn{1}{c}{\textbf{Gesture}} & \multicolumn{1}{c}{\textbf{RMSE}} & \multicolumn{1}{c}{\textbf{MAE}} & \multicolumn{1}{c}{\textbf{Avg $U_1$}}  \\ 
\hline\hline
\multirow{13}{*}{MSC}                & $\text{MSC}_{1}$                                & 4.101                             & 2.143                            & 0.069                                \\
                                     & $\text{MSC}_{2}$                                & 39.067                            & 12.283                           & 0.254                                \\
                                     & $\text{MSC}_{3}$                                & 69.123                            & 34.171                           & 0.468                                \\
                                     & $\text{MSC}_{4}$                                & 50.594                            & 24.116                           & 0.307                                \\
                                     & $\text{MSC}_{5}$                                & 52.428                            & 25.538                           & 0.370                                \\
                                     & $\text{MSC}_{6}$                                & 65.605                            & 31.718                           & 0.587                                \\
                                     & $\text{MSC}_{7}$                                & 23.665                            & 2.197                            & 0.030                                \\
                                     & $\text{MSC}_{8}$                                & 47.459                            & 20.070                           & 0.270                                \\
                                     & $\text{MSC}_{9}$                                & 55.631                            & 26.040                           & 0.313                                \\
                                     & $\text{MSC}_{10}$                               & 68.613                            & 47.524                           & 0.796                                \\
                                     & $\text{MSC}_{11}$                               & 79.191                            & 48.210                           & 0.841                                \\
                                     & $\text{MSC}_{12}$                               & 45.385                            & 5.860                            & 0.094                                \\
                                     & $\text{MSC}_{13}$                               & 68.059                            & 23.202                           & 0.290                                \\
\hline\hline
\end{tabular}
\end{table}



\bibliographystyle{IEEEtran}
\bibliography{refs}
\vspace{-10 mm}
\begin{IEEEbiography}[{\includegraphics[width=1in,height=1.25in,clip,keepaspectratio]{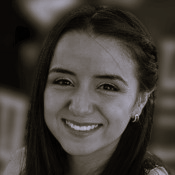}}]{Brenda Elizabeth Olivas-Padilla} received a B.S. degree in mechatronics engineering from Monterrey Institute of Technology and Higher Education, Mexico, in 2016, an M.S. degree in electronics engineering from the National Technological Institute of Mexico, Mexico, in 2018, and a Ph.D. degree in real-time computer science, robotics, systems and control from the Université PSL, France, in 2023. From 2018 to 2019, she was a Research Engineer at the Center of Robotics, Mines Paris, Université PSL, France, a Doctoral Student from 2019 to 2022, then a Post-Doctoral Researcher since 2023 at this same institution. Her research is focused on wearable sensing and movement modeling for the human movement analysis of expert artisans and operators from manufacturing industries. In addition, she has worked on two H2020 projects, where she contributed to the capturing, processing, and analyzing of technical movements for human learning and ergonomics. 
\end{IEEEbiography}
\vspace{-10 mm}
\begin{IEEEbiography}[{\includegraphics[width=1in,height=1.25in,clip,keepaspectratio]{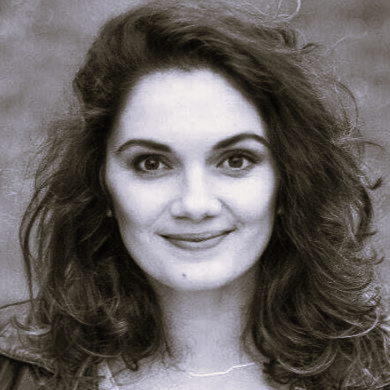}}]{Alina GLUSHKOVA} received her B.S. degree in information and communication from the University Paris 8, France, and two master degrees in management from the University of Paris Pantheon Assas and the University of Paris Dauphine, France, in 2012. She received her Ph.D. degree in applied informatics from the University of Macedonia, Greece, in 2016. Since 2020, she has worked as a Researcher at the Centre for Robotics at Mines Paris, where she co-coordinates the post-master AIMove – "Artificial Intelligence and Movement in Robotics and Interactive Systems." Her research focuses on human-computer interaction and, more specifically, on developing systems and algorithms that improve machines' perception and make them capable of analyzing and evaluating human movement with the aim of providing feedback that would guide the gestural/postural performance of the learner. Her work has been applied in the analysis and correction of the ergonomics of the professional gesture, as well as in the learning of the expert gesture.
\end{IEEEbiography}

\begin{IEEEbiography}[{\includegraphics[width=1in,height=1.25in,clip,keepaspectratio]{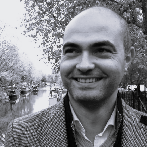}}]{Sotiris Manitsaris} is Deputy Director at the Centre for Robotics at Mines Paris, Université PSL. He holds a B.S. degree in applied mathematics from the Aristotle University of Thessaloniki, Greece, a double Master degree in local development from the University of Blaise-Pascal, France, and the Engineering School of the University of Thessaly, Greece, and a Ph.D. degree in artificial intelligence from the University of Macedonia, Greece. In addition, he did three post-doctoral research in biomedical engineering, human-robot collaboration, and movement-based interactive systems. He has been the principal investigator of a number of European public and industrial projects and the leader of the post-master AIMove – "Artificial Intelligence and Movement in Robotics and Interactive Systems." In 2020, he joined the International Advisory Board of the STOA Panel of the European Parliament, which puts a specific emphasis on the field of AI through its newly-established Centre for AI. His research interests put a special focus on human-centred artificial intelligence that consists of machine learning and pattern recognition concepts and methods that are applied to signals recorded from the human body and used as modalities for collaborating with intelligent machines.
\end{IEEEbiography}

\EOD

\end{document}